\documentclass{article}

\pdfoutput=1
\usepackage{arxiv}

\usepackage[utf8]{inputenc} 
\usepackage[T1]{fontenc}    
\usepackage{hyperref}       
\usepackage{url}            
\usepackage{booktabs}       
\usepackage{amsfonts}       
\usepackage{nicefrac}       
\usepackage{microtype}      
\usepackage{lipsum}		
\usepackage{graphicx}
\usepackage{doi}

\usepackage{amssymb,amsmath,latexsym}
\usepackage[english]{babel}
\usepackage{caption}
\usepackage{subcaption}
\usepackage{multirow}

\usepackage{algorithm2e}
\usepackage{setspace}

\newtheorem{rem}{Remark}
\newtheorem{defi}{Définition}
\newtheorem{theo}{Theorem}

\title{Predicting the Future is like Completing a Painting!\\
\begin{small}
Unthinkable Novel Method for Time-series Forecasting Arising from Philosophical Considerations
\end{small} }

\author{ {Nadir Maaroufi, Mehdi Najib, Mohamed Bakhouya}\\
	International University of Rabat\\
	College of Engineering and Architecture, TICLab, LERMA Lab \\
Sala Al Jadida 11000, Morocco\\
\texttt{\{nadir.maaroufi,mehdi.najib,mohamed.bakhouya\}@uir.ac.ma} \\
}
\date{}

\renewcommand{\headeright}{}
\renewcommand{\undertitle}{}
\renewcommand{\shorttitle}{\textit{Predicting the Future is like Completing a Painting!} }

\hypersetup{
pdftitle={A template for the arxiv style},
pdfsubject={q-bio.NC, q-bio.QM},
pdfauthor={David S.~Hippocampus, Elias D.~Striatum},
pdfkeywords={First keyword, Second keyword, More},
}

\begin{document}
\maketitle

\begin{abstract}
This article is an introductory work towards a larger research framework relative to \textit{Scientific Prediction}. It is a mixed between science and philosophy of science, therefore we can talk about \textit{Experimental Philosophy of Science}. As a first result, we introduce a new forecasting method based on image completion, named \textit{Forecasting Method by Image Inpainting (FM2I)}. In fact, time series forecasting is transformed into fully images- and signal-based processing procedures. After transforming a time series data into its corresponding image, the problem of \textit{data forecasting} becomes essentially a problem of \textit{image inpainting} problem, i.e., completing missing data in the image. An extensive experimental evaluation is conducted using a large dataset proposed by the well-known M3-competition. Results show that FM2I represents an efficient and robust tool for time series forecasting. It has achieved prominent results in terms of accuracy and outperforms the best M3 forecasting methods.    
\end{abstract}


\section{Introduction}
It is well known that good forecasts are of utmost importance in all scientific disciplines \cite{68}. However, the relevance of accurate forecasts has become increasingly palpable since the emergence of Big-Data as a new paradigm of scientific activities for many researchers \cite{8}. One of the suggested tracks was due to the distinguished statistician Leo Breiman when he wrote with reference to predictions: \textit{"If our goal as a field is to use data to solve problems, then we need to move away from exclusive dependence on data models and adopt a more diverse set of tools."}\cite{16}. Therefore, we believe that understanding the \textit{nature of prediction} can help creating novel forecasting methods. First of all, let us precise that forecasting is considered as a process of predicting the future according to past and present data. We emphasize that this terminology may vary according to each scientific discipline (see \cite{17}). 

This paper is not mainly concerned by \textit{philosophy of science}, but we want to share with the research community the methodology we have undertaken, starting by deeper philosophical considerations, with the aim is to elaborate a new forecasting method. More precisely, we first start by raising philosophical questions, which are related to the nature of prediction itself, and then translate these questions following a scientific methodology for being applied for real-sitting forecasting scenarios. We name this way of thinking the \textit{Experimental Philosophy of Science}. Thus, we will switch, in this article, between science and philosophy of science as much as required in order to better describe our approach and highlight obtained results.  

As a first result, we introduce a new time series forecasting method based on \textit{image completion}, we name it \textit{Forecasting Method by Image Inpainting (FM2I)}. It represents a performing and robust tool for Time Series (TS) forecasting. Basically, the main idea behind the FM2I method is as follows. We first transform a TS into an image and then complete it using adapted image inpainting techniques. The completed image is converted back to the TS in order to obtain forecast values. Our aim is to perform these tasks without usual pre-processing (e.g., seasonality, trends, stationarity) of the original TS. The process of transforming TS into images and from the inpainted images into the original TS, including forecast values, must be possible. 

In summary, the main general contributions of this work are two folds: \textit{i)} interlinking philosophy of science and science, named \textit{experimental philosophy of science}, \textit{ii)} an extensitivity testing framework and \textit{iii)} a novel forecasting method by image inpainting (FM2I). The remainder of this paper is structured as follows. Section 2 presents the formulation of the scientific prediction and philosophical problem. The scientific transformation of the problem is introduced in Section 3. In Section 4, materials and methods are described. The proposed FM2I algorithm is detailed in Section 5. Experimental results and comparison are described in Section 6. Scientific and philosophical discussions together with potential directions are given at the end of this document. An appendix is included in order to describe some notions and fundamentals related to signal theory.

\section{Scientific prediction and philosophical problem formulation}
It is well known in the philosophy of science field that \textit{Scientific Explanation} and \textit{Scientific Prediction} are one of the main goals of sciences \cite{1}, \cite{3} and \cite{4}. Nevertheless, philosophers of science have focused much more on developing a theory of explanation rather than introducing a prediction theory \cite{5}, \cite{2a} and \cite{2}. Indeed, several models of explanation were arised during the past half century. The pioneering and most famous models of explanation, so-called covering-law due to Carl Gustav Hempel and Paul Oppenheim, are DN (Deductive Nomological), DS (Deductive Statistical) and IS (Inductive Statistical) \cite{6}, \cite{7}. According to their models, we get a scientific explanation of an event/phenomenon $E$, which occurred (called \textit{Explanandum}) when $E$ can be deduced, induced or inferenced from the \textit{Explanans} ($I$,$L$), i.e., a set of initial conditions $I$ and a set of laws $L$. In other words, the explanandum $E$ must be a logical consequence of the explanans ($I$,$L$), which meet certain conditions \cite{6}. There are other subsequent alternative models of explaination; we can mention, for instance, the Causal \cite{11} or Unificationist \cite{12} model.

However, despite the extensive work in this area, still a logical prediction model is required to formalize the structure of predictive reasoning, see \cite{2a} and \cite{2}. One of the consequences is the weakening of the explanation theory itself \cite{5}. In order to overcome the difficulty of constructing a theory of prediction, early renowned works of Hempel and Oppenheim \cite{6} and \cite{7} introduced the controversial (\cite{9} and \cite{10}) \textit{Structural Identity Thesis} (also called \textit{Symmetry Thesis}) between explanation and prediction. The structural identity thesis assumes that explanation and prediction have the same logical form and are, somehow, equivalent. In other words, every adequate explanation is a potential prediction and every adequate prediction is a potential explanation. In particular, as it was recalled in \cite{7}, in the Hempel-Oppenheim model, predictions concern exclusively phenomena that occur in the future and, therefore, it seems that the difference between explanation and prediction is purely chronological or temporal. This structural identity thesis has been heavily criticized by many philosophers \cite{10}, \cite{9}, \cite{15}. 

Nowadays, there is no clear consensus concerning the definition of scientific prediction, see \cite{2a} and \cite{2}. Most particularly, the notion of scientific prediction itself is still ambiguous. Finding, however, an adequate and unified definition of prediction is challenging. It must be quite restrictive in order to exclude \textit{Divination}, since it is based on occultic and metaphysics process of predicting, for instance as in astrology. Moreover, it must be fairly too broad by including all forms arising from different scientific disciplines. Indeed, according to the definition of Hempel-Oppenheim's model, the evolutionary biology, palaeontology or the geology are automatically excluded, since these disciplines deal with the past, which means that they are exclusively explicative sciences. Furthermore, their model also excludes any prediction not obtained through scientific laws (see a counter-example in \cite{18}).

Actually, in current literature, a better understanding of the notion of scientific prediction appears in two major areas of the philosophy of science, \textit{Confirmation Theory} and \textit{Scientific Realism} debate see \cite{2b} and \cite{2}. Briefly, the two questions asked in these two fields are respectively : should a hypothesis or a theory be better confirmed when it generates accurate predictions \cite{21}? Are accurate predictions a reasonable criterion to assess a theory's maturity in order to affirm the reality of its assumed structures \cite{20}? These two philosophical domains reach the same problem related to scientific prediction and the need to redefine this fundamental notion is required. The article \cite{22} was the pioneering work, which explicitly makes the connection between these two philosophical areas and provides the current form of the issue \cite{2b}. We wish to emphasize that \cite{22} marks a certain break from the existing literature since it addresses a realistic predictive process from applied mathematics. Indeed, authors in \cite{22} analyze in a concrete way how existing regression analysis, from statistics theory, proceeds to predict, in order to make a powerful contribution to the philosophical debate.

In this work, we want to take the research a step further by philosophically formulating one of our questions concerning the nature of predictions. We will turn it into a scientific issue, and then formalize our own methodology, test it and then contribute to engaging both philosophical and scientific debate related to the prediction. That is what we mean by \textit{experimental philosophy of science}. We must stress that unlike the above-mentioned confirmation theory and scientific realism, our investigation is motivated by the prediction itself and not serving primarily another theory. Since, we are not yet able to formalize a definition of prediction, we need, however, a framework to better clarify our approach. Thus, we consider the definition given in \cite{23} or its extended version in \cite{2}. More precisely, the prediction is a result of a \textit{predictive process}. This later is considered as a series of inferences allowing setting, without measures or observations, the value of one or several variables (or the relationships between variables), variables are used here in a broad sense. However, this definition remains barely ambiguous since it is difficult to characterize a \textit{scientific predictive process} \cite{2}.  

Thus, we start by wondering the simple following question, which we call the \textit{nature of predictions question}: using the definition above, predictions are a result of a predictive process, then is this last possess a signature related to the own nature of the phenomenon, the event or the type of data we want to predict? In other words, if we dispose or establish a predictive process in a particular framework, can this process be potentially a predictive process in a completely different setting? More precisely, assume that we have an accurate predictive process arising from a theory or a model, concerning a specific scientific field, and able to predict a phenomenon $A$, is this process capable of predicting, in another scientific field, a phenomenon $B$? This question may seem to be trivial for theories, which are specifically concerned by prediction. For example, the statistical theory is able to predict a variety of different phenomena, since we can model them according to known tools or classes of statistics-based models. In our research program, since the aim is to exhibit new prediction methods, we want to go further by investigating "unthinkable" links between phenomena.           

After careful considerations, it appeared suitable to reformulate this question in relationship with the notion of \textit{Predictive Capacity}. First of all, we want to precise the difference between the latter and \textit{Predictive Power}, see \cite{2a} for more details. The predictive power refers to the ability to derive, from a theory, at least one testable prediction. It can be obtained \textit{a priori} by ensuring that the theory might have empirical consequences. While the predictive capacity can be used to evaluate the variety and the accuracy of the predictions generated by a theory, there are no ways, however, to assess it \textit{a priori} since the validity of these predictions have to be checked after their future realizations (or not) and that in relationship with the confirmation theory, as stated by \cite{2a} \cite{2b}. The same author gives two dimensions or two epistemic virtues of predictive capacity: the \textit{Intensivity}, which refers to the accuracy of predictions with regard to a particular phenomenon, while the \textit{Extensivity} refers to the ability to predict in a large variety of phenomena. Accordingly, we can formulate the following question: how can we test the extensivity of a theory's or a model's predictive capacity?

\section{Scientific transformation of the problem}

\begin{figure}[h!]
\centering
  \includegraphics[scale=0.45]{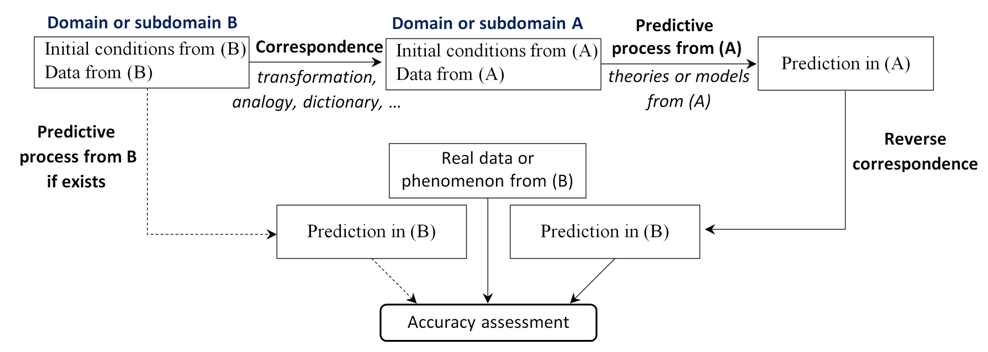} 
  \caption{Extensivity testing framework}
\label{Fig1}
\end{figure}

As described in the previous section, the problem is formulated in a very general way, but the power of this type of formulation is to generate potential research directions in the field of scientific prediction. More precisely, throughout an adequate \textit{correspondence} procedure, the purpose is to test the extensivity of a theory's or a model's predictive capacity from a specific scientific setting or domain $A$. Thus, we have to choose a predictive process, which is designed, established and used in $A$, then test it in the frame of another specific domain $B$. This \textit{correspondence} must enable switching, going back and forth, between the two domains or sub-domains. The results of this procedure are assessed in terms of accuracy according to real and/or existing predictions, which are, eventually, obtained by a predictive process from $B$. This procedure, named \textit{extensivity testing framework (ETF)}, can be formalized as depicted in Fig. \ref{Fig1}.

While this extensivity testing framework could certainly bring a first answer to the previous question, it will not obviously tackle the whole above-mentioned philosophical problem. Nevertheless, getting a conclusive test, means that we have found a new, unknown and "unthinkable" prediction method and thus reaching our objective announced in the introduction. Therefore, it will be crucial to narrow the scope by only focusing, for instance, on the prediction by \textit{Extrapolation}/\textit{Interpolation}. So, let us recall that in practice, extrapolation or interpolation are operations, which are mainly used for predicting the values of one or several variables, denoted $Y$, according to known values of one or several variables, denoted $X$. The unique difference between the extrapolation and interpolation is as follows: the latter is concerned by the variables, which are situated inside the experimental or the studied sampling area, while the former is related to those that are outside this area \cite{2a}. However, since the coming trend is not known, the extrapolation seems a difficult task than the interpolation. But, both the extrapolation and interpolation can be clearly considered as predictive processes according to the definition given in the previous section. 

After careful considerations, we have focused on the extrapolation by testing the predictive capacity of \textit{Image Completion} for performing \textit{Time Series Forecasts}. Basically, an image is a distributed amplitude of colors \cite{24}, it is a two-dimensional spatial pixels organization. Furthermore, an image possesses a specific structure \cite{25}, which is constituted, for example, of edges and texels; while a temporal series is a one-dimensional temporal sequence of successive values. As a matter of fact, there is a significant difference between these two data structures. Moreover, the image-based predictive process arises from image restoration techniques, which are designed for spatial patterns analysis and governed by \textit{Visual Logic} assessment as well, such as image quality \cite{26} and \textit{Psychovisual} quality metrics \cite{27}. Thus, there is no reason to expect that this image-based predictive process could work for temporal series forecasting, since it should be based on a \textit{Chronological Logic} by analysing past/present patterns. In other words, finding a way to draw past/present events and completing the painting, obtained by a purely pictorial reasoning, allows to 'see' the future!  

\begin{figure}[htb]
\begin{center}
\includegraphics[width=0.8\linewidth]{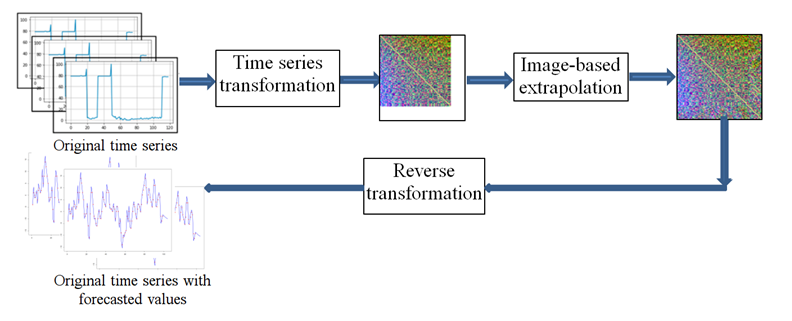}
\caption{A schematic view of the proposed fully integrated framework FM2I} 
\label{Fig2}
\end{center}
\end{figure}

Following the above-mentioned considerations, we have reached the required ingredients that allow developing our scientific and technical framework, which is considered as an instantiation of the proposed extensivity testing framework. It is basically a fully integrated modeling and processing framework for time-series forecasting, named FM2I (forecasting method by image inpainting), as depicted in Fig. \ref{Fig2}.

\section{Materials and methods}
The methodology presented in this section covers the entire process, from time series transformations into 2D image structure to image-based forecasting (inpainting), back to the 1D sequence representing the original times series including the forecast data. As shown in Fig. \ref{Fig2}, the FM2I approach is composed of three main processing phases: transformation of 1D sequence representing time series into two 2D image-based representation, image extrapolation by adapting inpainting methods, and reverse data techniques to transform a 2D image-based into 1D sequence to further represent the original time series together with forecast data. 

Regarding the first phase, the main question is how to transform 1D TS to 2D image-based structures by enhancing the initial information? We have investigated three main increasingly enhanced approaches. The first basic one, named \textit{naive approach}, considers the plot of the TS as an input image to the adapted inpainting method. In the second one, named \textit{enhanced approach}, we transform directly the TS into an augmented TS based-matrix using signal processing principles and tools. Based on this enhanced matrix, more adequate transformations are proposed in the deeper and final approach.   

The second phase is dedicated to image-based forecasting using adapted image inpainting methods. It is worth noting that these methods have been mainly proposed for filling in part of an image (restoration of damaged regions). Existing inpainting methods can be classified into two main categories \cite{60}: \textit{diffusion-based} approaches and \textit{examplar-based} approaches. Diffusion-based approaches have been mainly inspired by techniques from computational fluid dynamics. The pioneer work in this area was performed by Bertalmio et al. in \cite{61, 62} by using Navier-Strokes equation. Authors built an approach, at pixel level, by performing an analogy between the stream function in a 2D incompressible fluid and image intensity. As stated in \cite{60}, these approaches have provided excellent results for small inpainted regions, but they tend to introduce smooth effects in large regions. Exemplar-based approaches operate, however, at patch level by propagating information from known regions into missing ones. Unlike diffusion-based approaches, exemplar-based approaches are more adapted when inpainting large missing regions \cite{60}. 

In this work, we have adapted the exemplar-based approach, in particular, the patch-based method, which is basically proposed for completing missing (or deteriorated) parts of images. In fact, for TS forecasting, after transforming the TS into an image, the missing parts (i.e., unknown region) are considered as our forecast area. More precisely, the patch-based method is diverted from its basic function and then used for completing the area, which is related to the forecast TS values. It is worth noting that we are developing a modified Navier-Stokes based method. However, in this work, we choose to focus only on the adapted patch-based method.  

The rest of this section first presents the three TS transformation techniques (i.e., Naive, Enhanced, and Deeper) followed by a detailed FM2I algorithm together with issues we have tackled in order to maximize the transformation precision while considering a back and forth \textit{TS-image correspondence}.

\subsection{Naive approach}
\begin{figure}[!h]
\centering
\begin{subfigure}{.5\textwidth}
  \centering
  \includegraphics[width=.9\linewidth]{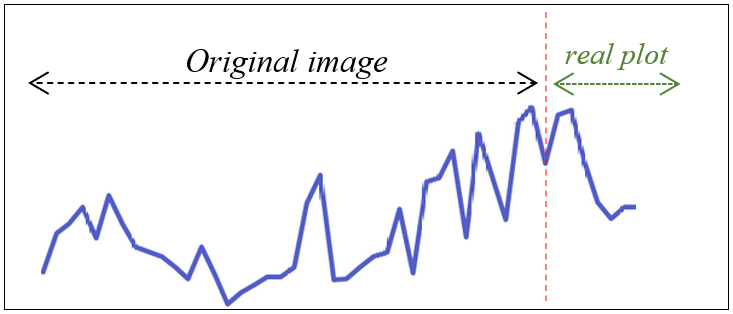}
  \caption{}
  \label{Fig3}
\end{subfigure}%
\begin{subfigure}{.5\textwidth}
  \centering
  \includegraphics[width=.9\linewidth]{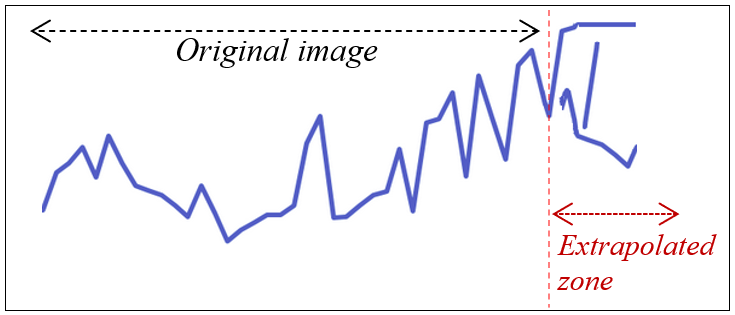}
  \caption{}
  \label{Fig4}
\end{subfigure}
\caption{Actual TS with a) real data, b) predicted data}
\label{Naive}
\end{figure}

In this approach, we have considered the plotted TS as a 2D image (Original image), as depicted in Fig. \ref{Naive}. The plotted TS represents CO2 values from the dataset described in \cite{44}. As shown in this figure, the forecast area (extrapolated zone) is added to the original image as a deteriorated part for being inpainted by the patch-based method. The latter considers the extrapolated area as a regular hole to be reconstructed or inpainted, i.e., filling the missing pixels. The adapted patch-based method is able to fill on it; however, it generates several forecasts with high errors compared to the original TS (i.e., real plot). In fact, it is difficult, even impossible, for the patch-based method to infer the forecast area. It is due to the fact that the plotted TS 2D image holds less rich information. Indeed, there are much more white pixels than those representing TS values. Moreover, we have noticed, from the obtained experimental results, that the patch-based method does not deal well with such representations. This constraint compelled us to find suitable and richer 2D representations of TS, as described in the next section. Our aim is to extend the original TS with richer information, which could lead to more image patterns' structures and features.

\subsection{Enhanced approach}
The aim of this enhanced approach is to generate the TS's corresponding matrix that could increase the information gain against such 1D TS representation. Basically, a time series $TS$, having a set of values indexed in a time order, can be represented by $TS=\{s_0,s_1,s_2,s_3,....s_n\}=(s_i)_{i=0,1,..n}$, where $s_0$ is the first value obtained at $t=0$ while $s_n$ represents the value obtained at $t=n$. We have investigated existing techniques, from the signal processing theory, in order to transform a 1D signal into a 2D matrix. More precisely, we have used the \textit {Time Autocorrelation Function} as a tool for 1D-to-2D transformations. Let us consider a TS as a \textit{random signal with a finite average power}, the aim is to represent the majority of real-life time series. We can consider a TS as a \textit{digital signal} resulting from the sampling of a random and permanent analog bounded signal $x(t)$ having a finite average power. In fact, let us consider $x(t)$ a signal having a finite average power, we can then define the temporal autocorrelation of this signal by the following formula, but more details are presented in the Appendix A: $$\Gamma(\tau)=\lim_{T\to +\infty}\frac{1}{T}\int^{\frac{T}{2}}_{-\frac{T}{2}} x(t)x(t-\tau)dt$$. 

It is worth noting that the function $\Gamma(\tau)$ is well defined. Indeed, by applying the Cauchy-Schwarz inequality taking into consideration that $x(t)$ is bounded, we can obtain $\Gamma(\tau)\leq P(x(t))<+\infty$. Actually, the maximum of the function $\Gamma$ is found at the origin, in fact, we have $\Gamma(0)=P(x(t))$. 

This autocorrelation function measures the degree of similarity between the signal and its offset version, so it is natural that this function is maximum for $\tau = 0$. $\Gamma(\tau)$ compares the signal to itself, it is, therefore, a temporal average of the product of the signal by itself shifted by a time $\tau$. Moreover, $\Gamma(\tau)$ measures the similarity degree or internal signal's dependency. Mathematically, this function can be seen in several ways; namely, as a scalar product between the signal and its offset. Thus, if $\Gamma(\tau) = 0$, it means that the signal and its shift of one step $\tau$ are completely uncorrelated, they are two orthogonal signals. We can also see $\Gamma$ as a sort of convolution of the signal with itself. This autocorrelation function does not only contain important information about the signal, but also its Fourier transform and then its spectrum. Indeed, this relationship manifests itself through the power spectral density (PSD). Thus, this autocorrelation function seems to contain valuable information regarding the properties of the signal $x(t)$. More details regarding the PSD relationship with the auto-correlation function are presented in the Appendix B.

For a digital signal $(x_n)_{n\in\mathbb{Z}}$, the discrete version of $\Gamma(\tau)$ can be written as follows:
$$\gamma(n)=\lim_{M\to +\infty}\frac{1}{M}\sum_{k=-\frac{M-1}{2}}^{\frac{M-1}{2}} x_kx_{k-n}$$. 

\noindent So, the autocorrelation formula for a TS, for $i\in [0,1,2...n]$, is $\gamma(i)=\frac{1}{n+1}\sum_{k=i}^{n}s_ks_{k-i}$. This allows defining a symmetric temporal autocorrelation matrix in the following form:

\begin{small}
\[ \gamma_{(i\in [0,1,..n])}= \begin{pmatrix}
\gamma(0) &\gamma(1) &\gamma(2)&\cdots &\cdots& \gamma(n) \\ 
\gamma(1) &\gamma(0)& \gamma(1)&\gamma(2) &\cdots & \gamma(n-1)\\
\gamma(2)&\gamma(1)& \ddots &\ddots&\vdots & \cdots\\

\vdots &\gamma(2)&\gamma(1) & \ddots&\ddots & \vdots\\
\vdots &\vdots&\gamma(2)& \ddots&\ddots & \gamma(1)\\
\gamma(n)&\gamma(n-1)& \cdots&\cdots & \gamma(1) & \gamma(0)\\ 

\end{pmatrix} 
\]
\end{small}

It is worth noting that, it is not only the symmetric temporal autocorrelation matrix (STAM) that can be generated, but also the STAM version by excluding its average. Indeed, any signal $x(t)$ can be written according to the form $x(t)=\bar{x}+y(t)$, where $\bar{x}$ is a constant, it is the mean value of the signal $x(t)$, and $y(t)$ is a signal of zero mean value. So, we can get the relation $\frac{dx(t)}{dt}=\frac{dy(t)}{dt}$. This will, therefore, allow using the (STAM) of the signal $y(t)$ with zero mean. More specifically, for a TS, differentiating it by doing $s_{i+1}-s_i$, provides a new TS, where the mean is eliminated before generating the STAM. Therefore, it will be easy to return back to the original TS afterwards, including the forecast area. More details about particular signals, which are stationary and ergodic are given in Appendix C.

\begin{figure}[!h]
\begin{center}
\includegraphics[width=0.9\linewidth]{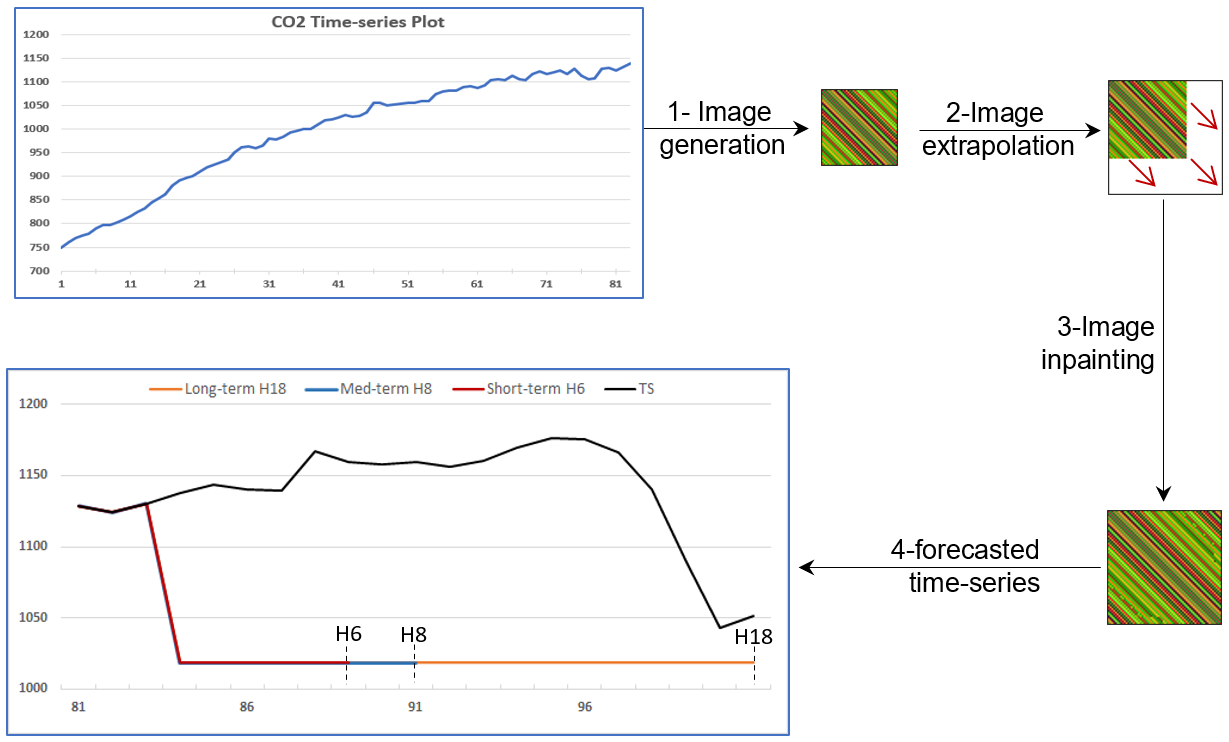}
\caption{Forecasting using the auto-correlation matrix} 
\label{Fig5}
\end{center}
\end{figure}

The auto-correlation function, which is used to generate the STAM, is used, in turn, for TS forecasting. In fact, this latter is applied to the original TS in order to generate its corresponding matrix, as depicted in Fig. \ref{Fig5} (image generation). Then, the image, including the area to be predicted, is generated (Delimited forecast area generation). The size of this area corresponds to the size of the forecast horizon. The adapted patch-based approach is then applied on this latter to extrapolate the forecast area (image inpainting). The final phase is the backtracking from the original image, including the forecast area (i.e., short, medium and long horizon), to its corresponding TS (forecast time-series). All these transformation techniques must take into consideration two major objective functions, maximizing the transformation precision (i.e., information loss induced by the transformation), minimizing the execution time, while considering a back and forth \textit{TS-image correspondence}. These techniques are described in Section 5.

In order to assess the effectiveness of this enhanced method, we have used the most common metrics, the MSE (Mean Squared Error), RMSE (Root Mean Square Error), MAE (Mean Absolute Error), the MAPE (Mean Absolute Percentage Error) and SMAPE (Symmetric Mean Absolute Percentage Error). The MSE is used to assess the quality of the estimation by computing the average squared difference between actual and estimated values. It is at all times positive, but when closer to zero the estimation is considered of high quality. Alike MSE, RMSE is used to measure the deviation of predicted values from actual values. It is computed as square root of average square deviation. The MAE is used to measure the average magnitude of the errors between forecast values and their corresponding actual ones. It is worth noting that the lower values of RMSE and MAE are better, i.e., the smaller an RMSE/MAE value is observed more the predicted values are closer. The MAPE measures the average of mean absolute percent errors, while sMAPE measures the average of symmetric mean absolute percent errors (\%). The metrics are formally described as follows: $MSE =  \frac{1}{n} \sum_{i=1}^n(F_{i} - A_{i})^2$, $RMSE =  \sqrt{ \sum_{i=1}^n(F_{i} - A_{i})^2 }$, $MAE =  \frac{\sum_{i=1}^n |F_{i} - A_{i}| }{n}$, $MAPE =  \frac{1}{n} \sum_{i=1}^n \vert \frac{F_{i} - A_{i}}{A_{i}}  \vert$, and $SMAPE =  \frac{2}{n} \sum_{i=1}^n  \frac{|F_{i} - A_{i}|}{|F_{i}| + |A_{i}|}$, where $n$ is the forecast horizon (n-step-ahead), $F_{i}$ are forecast values, which are produced at time $i$ by the established forecasting algorithm, whereas $A_{i}$ is the value, which is actually observed at time $i$.

\begin{table}[!h]
 
\scriptsize
\begin{center}
\begin{tabular}{|l|c|c|c|} 
   \hline
\textbf{Metrics} & \textbf{Short-Horizon} & \textbf{Med-Horizon} & \textbf{Long-Horizon}  \\
\hline
MSE              & 16864.50               & 17546.19             & 16375.84               \\
\hline
RMSE             & 129.86                 & 132.46               & 127.96                 \\
\hline
MAE              & 129.37                 & 132.02               & 122.05                 \\
\hline
MAPE             & 11.26\%                & 11.47\%              & 10.59\%                \\
\hline
sMAPE            & 11.94\%                & 12.17\%              & 11.24\%                \\
\hline
\end{tabular}
\end{center}
\caption{Performance metrics for short, medium and long horizon forecasting} 
\label{Metrics-Deep}
\end{table}

Table \ref{Metrics-Deep} shows the MSE, RMSE, MAE, MAPE and the sMAPE for short (6-step-ahead), medium (8-step-ahead) and long (18-step-ahead) forecasting. The obtained results show similar patterns, like the naive approach, but with high errors for all horizons, short, medium and long. More precisely, this enhanced method is able to generate only constant forecast values, which is an obvious behaviour since they show homogeneous patterns (i.e., diagonal lines form). In fact, the adapted patch-based method performs well by forecasting data, but the auto-correlation matrix is not suitable in this case. Moreover, despite the fact that we have used one of the best patch-based methods, its behaviour is mainly related to the image's patterns, which are related to the relevance of the selected TS-image transformations. This boils down to the fact that the information is too concentrated and needs to be broken down. The next section will introduce deeper transformation techniques including new representations of original TS.

\subsection{Deeper approach} 
\begin{figure}[!h]
\centering
\begin{subfigure}{.5\textwidth}
  \centering
  \includegraphics[width=.5\linewidth]{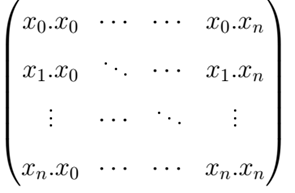}
  \caption{MAC}
  \label{MAT-MAC}
\end{subfigure}%
\begin{subfigure}{.5\textwidth}
  \centering
  \includegraphics[width=.7\linewidth]{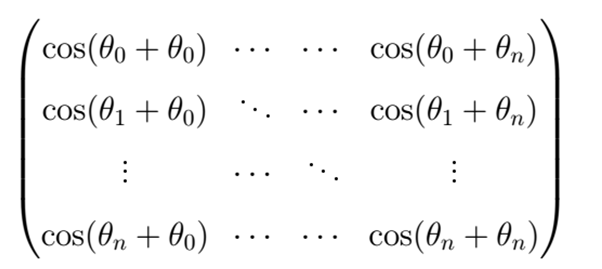}
  \caption{GASF}
  \label{MAT-GASF}
\end{subfigure}
\newline
\begin{subfigure}{.5\textwidth}
  \centering
  \includegraphics[width=.6\linewidth]{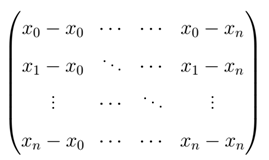}
  \caption{RPM}
  \label{MAT-RPM}
\end{subfigure}%
\begin{subfigure}{.5\textwidth}
  \centering
  \includegraphics[width=.7\linewidth]{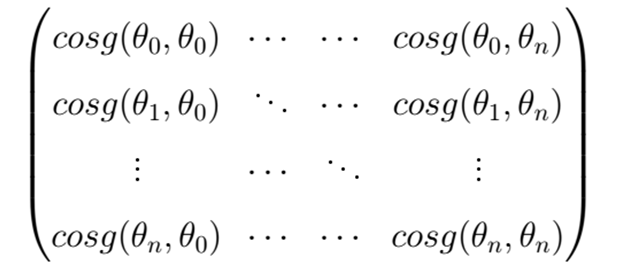}
  \caption{GC}
  \label{MAT-GC}
\end{subfigure}

\caption{Matrix representations of TS} 
\label{MAT}
\end{figure}

We have first investigated a new matrix representation, with the aim to enrich data through self-signal autocorrelation, called modified auto-correlation (MAC), by combining all possible offsets. This matrix can be performed by transforming the auto-correlation matrix into another one (MAC), for instance, by separating the sums, which are seem represented on the diagonal and the sub diagonals. We then obtain the matrix, depicted in Fig. \ref{MAT-MAC}, in which the sum of the diagonal and each diagonal makes the $\gamma(i)\times (N+1)$. This structure leads to a gramian matrix (GM).    

It is worth noting that in \cite{56} authors have reached similar results (Gramian Angular Field, GAF), by encoding TS (1D cartesian coordinates) as images (2D polar coordinates). In fact, images are represented by GAF in which each element is computed by the trigonometric sum between different time intervals. Furthermore, they have presented other techniques to allow transforming 1D sequence time series data into image-based representation for TS classification using traditional machine learning approaches. In fact, after transforming time series data into images, conventional neuronal approaches are used for classification purposes, e.g., learning features and identifying structure in TS. More precisely, instead of using TS data as input, these approaches use its corresponding encoded images (e.g., traditional image recognition). For instance, we can cite the gramian angular summation field (GASF), Gramian angular differenced field (GADF), markov transition field (MTF). Similarly, authors in \cite{58} present another technique, called relative position matrix (RPM), for transforming 1D TS into image-based representations (see Fig. \ref{MAT-RPM}). They mainly combined the RPM and CNN (Convolutional Neural Network) in an unified framework in order to enhance the accuracy of the time series classification. Reported results show the effectiveness of the proposed approaches against the state-of-art classification techniques. 

To the best of our knowledge, all research works, which use image-based representations of TS, are dedicated to classification. However, only the work presented in \cite{59}\cite{63} dealt with time series forecasting. However, in \cite{59}\cite{63} authors used the same principle adapted for classification purposes, i.e., from 1D vector into 2D representation, in the form of image, to data classification using conventional neuronal approaches. Furthermore, they have used an image-based representation similar to our above-mentioned naive representation. For instance, as described in \cite{59}, the TS (i.e., the daily load data) is transformed into a graphical representation instead of a matrix-based representation. A dual-branch deep convolutional network is then fed by the generated graphical (i.e. plotted) representation in order to extract features for clustering purposes. 

After a detailed study of the above-mentioned representations, we have selected GASF and RPM, which allow easily a return back to the 1D (i.e., 1D-2D and 2D-1D) representation. For the GASF transformation matrix, authors, in \cite{46}, presented the transformation as a transition to polar coordinates, assuming that $\theta_0=\arccos(x_0),\theta_1=\arccos(x_1),\theta_2=\arccos(x_2).....\theta_n=\arccos(x_n)$. More specifically, if $\theta_0=\arccos(x_0),\theta_1=\arccos(x_1),\theta_2=\arccos(x_).....\theta_n=\arccos(x_n)$, the matrix can be written as depicted in Fig. \ref{MAT-GASF}. According to the trigonometry formulas, the following expression, $\cos(\theta_i+\theta_j)=x_i.x_j-\sqrt{(1-x_i^2)}\sqrt{1-x_j^2}$, can be easily obtained. Unlike MAC representation, the GASF uses a modified scalar product representation. 

Alike the GASF representation, we have introduced another matrix, named Generalized cosine (GC). After rescaling the TS values to be within unified interval (e.g., $[-1,1]$), if we denote by $cosg$ the generalized cosine function, as introduced in \cite{43}, the two-variables function is defined by $cosg(\theta_i,\theta_j)=\frac{\sin(\theta_i)}{\sin(\theta_i+\theta_j)}$, and therefore, the corresponding GC matrix can be generated, as depicted in Fig. \ref{MAT-GC}. Using the formulas of trigonometry we can get: $cosg(\theta_i,\theta_j)=\frac{\sqrt{(1-x_i^2)}}{x_i\sqrt{1-x_j^2}+x_j\sqrt{1-x_i^2}}$, and $cosg(\theta_i,\theta_i)=\frac{1}{2x_i}$. As shown in Fig. \ref{MAT-GC}, the matrix is not symmetric. Therefore, in order to reinforce TS representation patterns and features, we have proposed two symmetric forms of the basic GC, GCS1 and GCS2 as follows: $GSS1=\frac{1}{2}(GS+^tGS)$ where $^tA$ is the transpose of matrix $A$ and $GSS2(i,j)=GS(j,i)$ for $i<j$. 

\begin{figure}[!h]
\centering
\begin{subfigure}{.16\textwidth}
  \centering
  \includegraphics[width=1\linewidth]{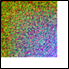}
  \includegraphics[width=1\linewidth]{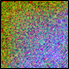}
  \includegraphics[width=1\linewidth]{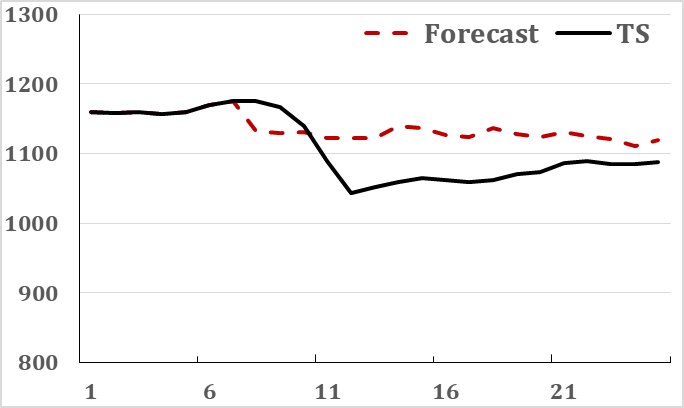}
  \caption{MAC}
  \label{For-SP}
\end{subfigure}
\begin{subfigure}{.16\textwidth}
  \centering
  \includegraphics[width=1\linewidth]{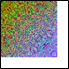}
  \includegraphics[width=1\linewidth]{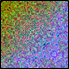} 
  \includegraphics[width=1\linewidth]{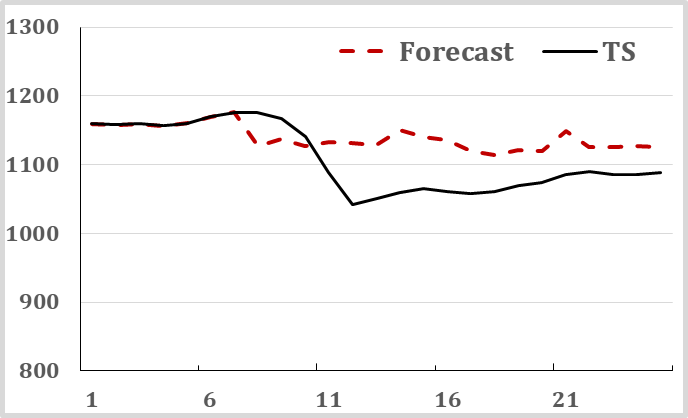} 
  \caption{GASF}
  \label{For-GASF}
\end{subfigure}
\begin{subfigure}{.16\textwidth}
  \centering
  \includegraphics[width=1\linewidth]{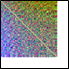}
  \includegraphics[width=1\linewidth]{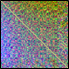}
  \includegraphics[width=1\linewidth]{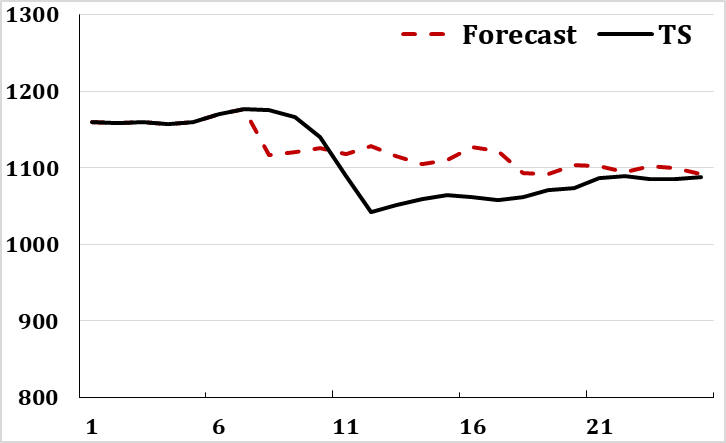}
  \caption{RPM}
  \label{For-RPM}
\end{subfigure}
\begin{subfigure}{.16\textwidth}
  \centering
  \includegraphics[width=1\linewidth]{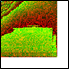}
  \includegraphics[width=1\linewidth]{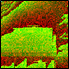}
   \includegraphics[width=1\linewidth]{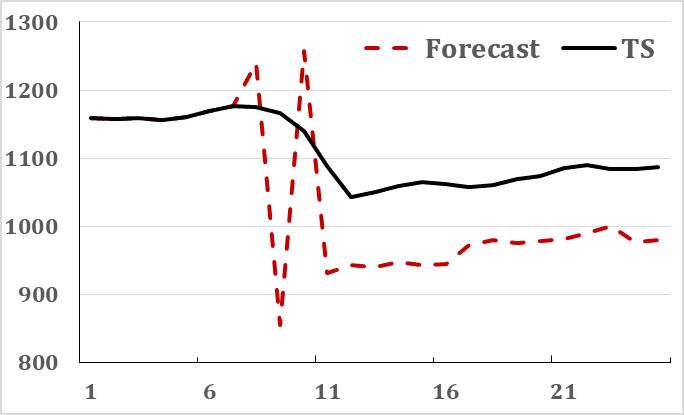}
  \caption{GC}
  \label{For-GC}
\end{subfigure}
\begin{subfigure}{.16\textwidth}
  \centering
  \includegraphics[width=1\linewidth]{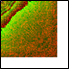}
  \includegraphics[width=1\linewidth]{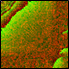}
  \includegraphics[width=1\linewidth]{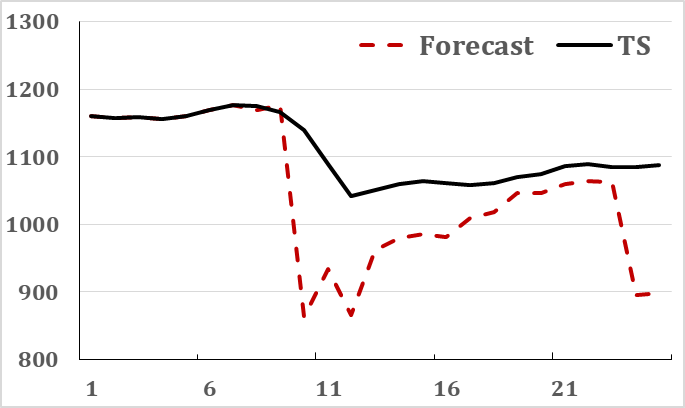}
  \caption{GCS1}
  \label{For-GC-Sym1}
\end{subfigure}
\begin{subfigure}{.16\textwidth}
  \centering
  \includegraphics[width=1\linewidth]{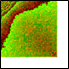}
  \includegraphics[width=1\linewidth]{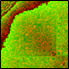}
  \includegraphics[width=1\linewidth]{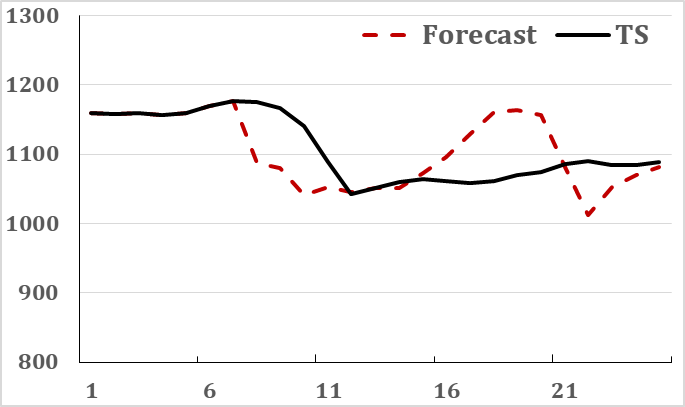}
  \caption{GCS2}
  \label{For-GS-Sym2}
\end{subfigure}

\caption{Image-based representation: without and with forecast area}
\label{MAT-FORCAST}
\end{figure}

In order to evaluate the efficiency of these matrix representations for TS forecasting, we have computed their accuracy using a set of TS (3003 time series) from M3 competition \cite{45}. Fig.\ref{MAT-FORCAST} represents an example of a given TS using the above-mentioned image-based representations. Extensive experiments have been conducted in order to compute the MSE, RMSE, MAE, MAPE and the sMAPE of selected TS (i.e., TS1, TS3, TS4, TS11, TS20, TS24). Obtained results are reported in Table \ref{FORTAB18} to show whether these matrices affect the forecasting accuracy. As shown in this table, none of them outperforms for all considered TS.   

\begin{table}[]
\centering
\scriptsize
\begin{tabular}{|l|c|c|c|c|c|c|r|r|r|r|r|r|}
\hline
                    & \multicolumn{6}{|c|}{\textbf{Matrices ranking}}                                                                                                                                                                              & \multicolumn{6}{|c|}{\textbf{sMAPE\%}}                                                                                                                                                                                                               \\ \hline
                    & \textbf{TS1}                       & \textbf{TS3}                       & \textbf{TS4}                       & \textbf{TS11}                      & \textbf{TS20}                      & \textbf{TS24}                      & \multicolumn{1}{|c|}{\textbf{TS1}}     & \multicolumn{1}{|c|}{\textbf{TS3}}      & \multicolumn{1}{|c|}{\textbf{TS4}}      & \multicolumn{1}{|c|}{\textbf{TS11}}     & \multicolumn{1}{|c|}{\textbf{TS20}}      & \multicolumn{1}{|c|}{\textbf{TS24}}      \\ \hline
GASF     & 5                                  & 2                                  & 4                                  & 2                                  & \textbf{1} & 2                                  & 9,984                                 & 6,563                                  & 9,285                                  & 4,36                                   & \textbf{11,271} & 4,335                                  \\ \hline
GC     & 3                                  & 5                                  & 5                                  & 3                                  & 5                                  & \textbf{1} & 5,255                                 & 13,413                                 & 9,566                                  & 4,471                                  & 25,871                                  & \textbf{4,242} \\ \hline
GCS1 & 4                                  & 4                                  & \textbf{1} & 5                                  & 4                                  & 4                                  & 8,225                                 & 12,258                                 & \textbf{4,415} & 6,063                                  & 25,069                                  & 4,898                                  \\ \hline
GCS2 & 6                                  & 6                                  & 2                                  & \textbf{1} & 6                                  & 3                                  & 14,328                                & 13,78                                  & 8,465                                  & \textbf{3,707} & 27,705                                  & 4,353                                  \\ \hline
MAC       & \textbf{1} & 3                                  & 6                                  & 4                                  & 2                                  & 5                                  & \textbf{1,18} & 10,459                                 & 10,865                                 & 5,321                                  & 12,648                                  & 5,946                                  \\ \hline
RPM     & 2                                  & \textbf{1} & 3                                  & 6                                  & 3                                  & 6                                  & 1,426                                 & \textbf{5,082} & 8,995                                  & 6,482                                  & 12,959                                  & 6,44                                   \\ \hline
\end{tabular}

\caption{Matrices ranking for various time-series: short-term forecasting}
\label{FORTAB18}
\end{table}

\section{FM2I algorithm}

\begin{figure}[!h]
\centering
\includegraphics[width=0.9\linewidth]{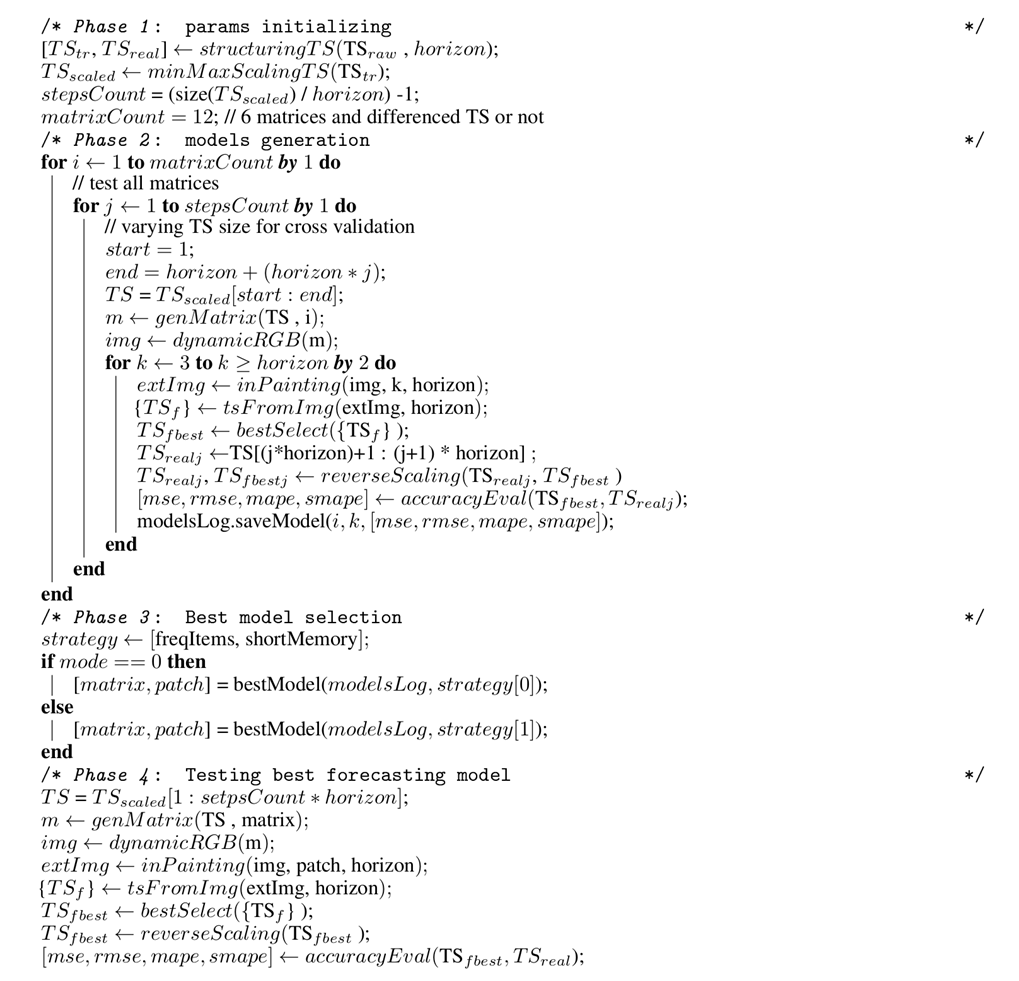}
\caption{FM2I Algorithm}
\label{Algo}
\end{figure}

The FM2I algorithm is structured into four main phases, as depicted in Fig. \ref{Algo}. The \textit{first phase} includes all steps that are required for preparing TS. The first step splits TS into two parts, training (i.e., parameters tuning) and testing. The testing part represents the forecasting values according to the horizon's sizes. It is mainly used once the best forecasting model is specified. It is worth noting that since the above-mentioned matrix transformations can be considered as a correlation between every two values of the TS, both TS parts need to be rescaled in order to have the same data encoding (i.e., within the same data range or same data magnitude). However, rescaling the TS values, for instance to $[-1,1]$, represents an issue for some matrices (e.g., GASF). This could happen mainly when bringing matrices' values back to the original TS. As a solution, we paid more attention regarding each transformation in order to select the most suitable rescale. For instance, for the GASF, we have rescaled the original TS to $[0,1]$ instead of $[-1,1]$, in order to allow the transformation back and forth, i.e., between the original TS to its corresponding image.

The \textit{second phase} consists of an exhaustive testing of various forecasting models through a grid-search approach. In fact, we test all possible combinations enriched by a progressive exploration of the studied TS. Consequently, the matrices (2D representation) representing TS are built on an incremental way, so as to be used to select the best models for each sub-part of the TS. The 2D representation of each TS has to be generated as an image. Thus, techniques for image encoding (RGB), from the generated matrix, have to be applied. We first start by a technique proposed in \cite{56}, which uses a (RGB) scale-based solution in order to transform TS values into image encoding values. This solution, despite its encoding capabilities, shows its limits when returning back to the original TS. In order to tackle this issue, we have first introduced a \textit{dictionary-based technique}. It is a RGB dictionary, which is used to establish a static mapping of each numerical value, from the scaled matrix into a unique three values (RGB), required for colored pixel creation. This solution allows image encoding without precision loss; albeit, it is a time consuming task as it requires numerous research requests in a high dimension dictionary of $256^{3}$ features. The second technique generates a RGB dictionary with minimum features representing only the numerical values existing in the scaled matrix. Though, this solution does improve the execution time, it restricts the domain of the forecast values to previous values existing in the scaled matrix and hence hinders the generation of new values. This issue represents a huge information loss. The third technique generates a gray shaded image based on a static dictionary with only 256 values representing the possible gray levels. This is quite similar to the second technique; it reduces drastically the image generation time, however, it induces an important information loss. The fourth technique, named, \textit{dynamic image encoding}, uses a bijective function allowing encoding the scaled matrix values into three values RGB ($256^{3}$). Moreover, it supports the unique generation of RGB code for the new forecast values while reducing the image generation time. 

The generated images are then extrapolated and inpainted using the adapted patch-based algorithm. Once images are inpainted, the reverse process is applied in order to both extract the forecast values from the image and reverse all the scaling steps, in adequacy with the applied matrices. The generated models are sorted on based on their accuracy (i.e., sMAPE) and are saved into a log file. 

The third phase exploits the log file as the basic Dataset to be mined in order to extract the best model for being used for TS forecasting. For this purpose, two strategies are proposed: the first one sorts all the models according to their performance and then applies frequent items search in order to extract the most frequent configuration based on the progressive models generation. The second strategy, named the short memory, aims to reduce the TS size used for image generation and seeks to extract the frequent items based on the last exploration of the TS. Both strategies are tested and could be alternated on a case by case basis. 

The fourth phase generates forecasts based on the best model configuration specified in the previous phase. The forecast is executed through an image generation according to the best matrix. This image is extrapolated and inpainted according to the best patch size. The forecasts are extracted from the image and rescaled to rescind the scaling and differencing changes. Once the forecasts are generated they are compared to the $TS_{test}$ part, representing the real values to forecast, extracted at the parameters initializing step. 

In summary, the original TS is first scaled to $[-1,1]$ for equalizing its initial values taking into consideration the possible TS upper and lower bounds. The scaled TS is then used to generate its corresponding matrix by, eventually rescaling again its values to be within the interval $[0,1]$. This step is related to the matrix used with the aim to make easy the back-and-forth ($B\& F$) transformation process. The rescaled matrix is used to generate its corresponding image using the above-mentioned techniques (e.ge., dynamic image encoding). Regarding image extrapolation, the patch-based algorithm, which was proposed mainly for image inpainting in order to fill damaged images, is adapted and applied for image extrapolation. In fact, the patch-based algorithm is enhanced for image extrapolation from the borders with a partial neighborhood, which does not surround the area of the image to be completed as conventionally applied in a given image inpainting field. The reverse techniques are then applied to get forecast values and compute the accuracy metrics.

\section{Experimental results}
In order to assess the effectiveness of the FM2I, we have used the dataset from M3 competition, which is considered as the main driver for developing and comparing forecasting methods. The M3 dataset includes 3003 TS while involving more than 30 forecasting methods. TS are categorized according to their types (e.g., industry, finance) and time intervals (e.g., Monthly, quarterly, yearly). In other words, for short-, medium-, and long-term forecasting, 645 yearly TS, 756 quarterly TS, and 1428 yearly TS are used respectively. The TS, which are classified as "Other", are an additional set, which is dedicated for medium-term forecasting. Table \ref{DS} summarizes the number of TS from each category and according to the period (e.g., Monthly, quarterly, yearly). So far, two forecasting horizons are considered, 8 for quarterly and 6 for yearly. All forecasts are available in order to assess the accuracy of new forecasting methods against state-of-the-art ones. It is worth noting that we are still conducting experiments of the FM2I for long horizon using monthly data (forecasting horizon, 18). Preliminary results are promising, but the whole results will appear in the published version of this work. 

In our experiments, we have used the forecasts that are made available by participants of the M3 competition. We also selected the methods, which are featured from the competition, and achieved great performance. In fact, we have considered the following top methods, from M3 competition \cite{45}, in order to assess the performance of FM2I for TS forecasting: ARARMA, Theta, ForecastPro. As stated in \cite{45} the Theta method performs extremely well in almost all considered TS and forecasting horizons. ForecastPro was also considered as a new method, which performs well in a similar way to Theta. ARARMA is a variant of the ARIMA method, which is considered the most sophisticated statistical-based forecasting method. We have also included the Naive method as a baseline, as also described in M3 competition, in order to assess the performance of FM2I methods against this benchmark. 

\begin{figure}[!h]
\centering
\includegraphics[width=0.5\linewidth]{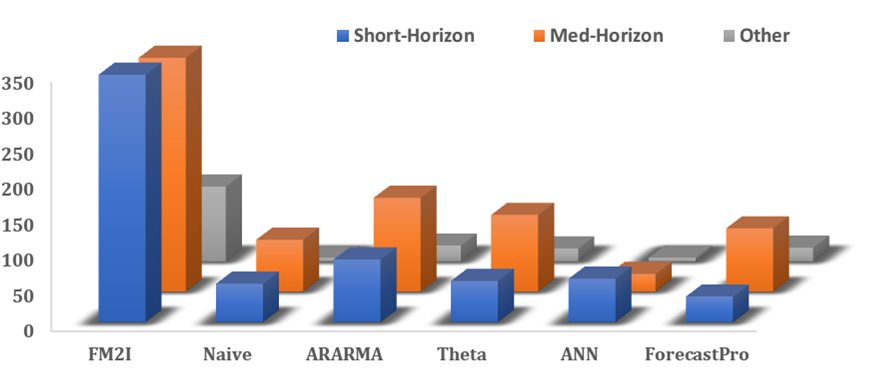}
\caption{The number of TS in which the methods show the highest accuracy}
\label{RANKFIG}
\end{figure}

Table \ref{SHORTFOR} presents the comparison, in terms of forecasting accuracy, of the FM2I against the above-mentioned methods. The accuracy is assessed in terms of MAE, MSE, RMSE, MASE and sMAPE. As shown in this Table, the FM2I outperforms all methods in all considered metrics. Similar behaviour is shown for medium term- (Table \ref{MEDIUMFOR}), as well for TS, classified as "Other" (Table \ref{OTHERFOR}). Similarly, we have computed the percentage and the number of time these methods are ranked best, as depicted in Table \ref{RANKTABLE}. As shown, the FM2I outperforms, with a large margin, the Naive, Theta, ForecastPro, ARIMA, and ANN approaches. Fig. \ref{RANKFIG} shows clearly the number of TS in which these methods are ranked the best by providing awesome accuracy. The FM2I performs exceptionally well and is ranked the best for the majority of TS.

\begin{table}[!h]
\scriptsize
\centering
\begin{tabular}{|l|l|l|l|l|l|l|l|}
\hline
\multicolumn{8}{|c|}{\textbf{Types of Time Series Data }} \\ 
\hline
\textbf{Interval}  & \textbf{Micro} & \textbf{Industry} & \textbf{Macro} & \textbf{Finance} & \textbf{Demog} & \textbf{Other} & \textbf{Total}  \\
\hline
\textbf{Yearly}    & 146            & 102               & 83             & 58               & 245            & 11             & 645             \\
\hline
\textbf{Quarterly} & 204            & 83                & 336            & 76               & 57             & 0              & 756             \\
\hline
\textbf{Monthly}   & 474            & 334               & 312            & 145              & 111            & 52             & 1428            \\
\hline
\textbf{Other}     & 4              & 0                 & 0              & 29               & 0              & 141            & 174             \\
\hline
\textbf{Total}     & 828            & 519               & 731            & 308              & 413            & 204            & 3003            \\
\hline
\end{tabular}
\caption{M3-Competition TS: the dataset used for FM2I evaluation}
\label{DS}
\end{table}

 \begin{table}[!h]
\scriptsize
\centering
\begin{tabular}{|l|r|r|r|r|r|r|r|r|r|r|}
\hline
\multicolumn{1}{|c|}{\textbf{Method}} & \multicolumn{5}{|c|}{\textbf{Average errors}} & \multicolumn{5}{|c|}{\textbf{Rank across all methods}} \\ \hline
                                     & MSE $.10^3$       & MAE       & RMSE         & MAPE $\%$ & sMAPE $\%$ & MAE      & MSE       & RMSE     & MAPE $\%$     & sMAPE  $\%$    \\ \hline
FM2I                                 & 1162,68      & 725,84       & 581,79      & 12.77    & 10.22     & 1        & 1         & 1         & 1       & 1          \\ \hline
Theta                                & 6626.00      & 1091.46      & 1252.70     & 22.58    & 16.97     & 4        & 4         & 4         & 5       & 2          \\ \hline
ForecastPro                          & 10706.26     & 1176.78      & 1354.30     & 22.23    & 17.27     & 5        & 5         & 5         & 4       & 3          \\ \hline
Naive                                & 2732.26      & 1025.84      & 1178.58     & 20.88    & 17.88     & 2        & 2         & 2         & 2       & 4          \\ \hline
ARARMA                               & 70250.21     & 1598.18      & 1896.22     & 26.73    & 18.35     & 6        & 6         & 6         & 6       & 5          \\ \hline
ANN                                  & 3201.64      & 1049.43      & 1215.62     & 21.83    & 18.56     & 3        & 3         & 3         & 3       & 6           \\ \hline
\end{tabular}
\caption{sMAPE and ranks of error for short-term horizon: 645 yearly TS}
\label{SHORTFOR}
\end{table}

\begin{table}[!h]
\scriptsize
\centering
\begin{tabular}{|l|r|r|r|r|r|r|r|r|r|r|}
\hline
\multicolumn{1}{|c|}{\textbf{Method}} & \multicolumn{5}{|c|}{\textbf{Average errors}} & \multicolumn{5}{|c|}{\textbf{Rank across all methods}} \\ \hline
                        & MSE $.10^3$  & MAE         & RMSE       & MAPE $\%$ & sMAPE $\%$ & MAE         & MSE   & RMSE       & MAPE $\%$     & sMAPE  $\%$    \\ \hline
FM2I      		& 524,38      & 441,82     & 350,97    & 9,27     & 6,66   & 1        & 1     & 1          & 1          & 1            \\ \hline
Theta                   & 850.32      & 475,41     & 557,23    & 11,67    & 8,95                                   & 2        & 2     & 2          & 2          & 2            \\ \hline
ForecastPro             & 1130.45     & 528,18     & 615,92    & 12,94    & 9,81                                   & 4        & 5     & 4          & 6          & 3            \\ \hline
Naive                   & 1026.69     & 523,73     & 611,44    & 12,38    & 9,95                                   & 3        & 4     & 3          & 3          & 4          \\ \hline
ANN                     & 995.24      & 534,46     & 642,69    & 12,81    & 10,15                                  & 5        & 3     & 6          & 4          & 5            \\ \hline
ARARMA                  & 1358.64     & 549,01     & 637,41    & 12,88    & 10,18                                  & 6        & 6     & 5          & 5          & 6          \\ \hline                          
\end{tabular}
\caption{sMAPE and ranks of error for med-term horizon: 756 quarterly TS}
\label{MEDIUMFOR}
\end{table}

\begin{table}[!h]
\centering
\scriptsize
\begin{tabular}{|l|r|r|r|r|r|r|r|r|r|r|} 
\hline
\multicolumn{1}{|c|}{\textbf{Method}} & \multicolumn{5}{c|}{\textbf{Average error }}                                                                                             & \multicolumn{5}{c|}{\textbf{Rank across all methods }}                                                                                    \\ 
\hline
                                      & \multicolumn{1}{l|}{MSE $.10^3$ } & \multicolumn{1}{l|}{RMSE} & \multicolumn{1}{l|}{MAE} & \multicolumn{1}{l|}{MAPE $ \%$} & \multicolumn{1}{l|}{sMAPE $ \%$} & \multicolumn{1}{l|}{MSE} & \multicolumn{1}{l|}{RMSE} & \multicolumn{1}{l|}{MAE} & \multicolumn{1}{l|}{MAPE $ \%$} & \multicolumn{1}{l|}{sMAPE $ \%$}  \\ 
\hline
FM2I                                  & 85.27                   & 144,54                   & 110,59                  & 2.66                     & 2.55                      & 1                        & 1                         & 1                        & 1                         & 1                          \\ 
\hline
ARARMA                                & 140.02                  & 213.53                   & 186.04                  & 4.67                     & 4.38                      & 2                        & 2                         & 2                        & 2                         & 2                           \\ 
\hline
Theta                                 & 208.93                  & 223.98                   & 197.11                  & 4.87                     & 4.40                      & 3                        & 3                         & 3                        & 3                         & 3                           \\ 
\hline
ForecastPro                           & 222.49                  & 235.21                   & 204.94                  & 5.10                     & 4.60                      & 5                        & 4                         & 4                        & 4                         & 4                           \\ 
\hline
ANN                                   & 214.77                  & 279.10                   & 226.83                  & 5.75                     & 5.23                      & 4                        & 5                         & 5                        & 5                         & 5                           \\ 
\hline
Naive                                 & 278.35                  & 309.88                   & 278.43                  & 7.02                     & 6.30                      & 6                        & 6                         & 6                        & 6                         & 6                           \\
\hline
\end{tabular}
\caption{sMAPE and ranks of error for mid-term horizon: other 174 TS}
\label{OTHERFOR}
\end{table}

\begin{table}[!h]
\centering
\scriptsize
\begin{tabular}{|l|l|l|l|l|l|l|l|}
\cline{3-8}
\multicolumn{1}{l}{}   &                      & \textbf{FM2I} & \textbf{Naive} & \textbf{ARARMA} & \textbf{Theta} & \textbf{ANN} & \textbf{ForecastPro}  \\
\hline
\textbf{Short-Horizon} & times ranked best    & 348           & 54             & 88              & 58             & 61           & 36                    \\
\hline
                       & \% ranked best model & 53,95\%       & 8,37\%         & 13,64\%         & 8,99\%         & 9,46\%       & 5,58\%                \\
\hline
\textbf{Med-Horizon}   & times ranked best    & 329           & 73             & 132             & 108            & 25           & 89                    \\
\hline
                       & \% ranked best model & 43,52\%       & 9,66\%         & 17,46\%         & 14,29\%        & 3,31\%       & 11,77\%               \\
\hline
\textbf{Other}         & times ranked best    & 105           & 5              & 22              & 18             & 5            & 19                    \\
\hline
                       & \% ranked best model & 60,34\%       & 2,87\%         & 12,64\%         & 10,34\%        & 2,87\%       & 10,92\%               \\
\hline
\end{tabular}
\caption{FM2I rank against other methods}
\label{RANKTABLE}
\end{table}

\begin{rem}[From the extensivity to the intensivity]
We have started this work wishing to find a new forecasting method by testing the extensivity of image inpainting methods. Our aim was to create new methods for TS forecasting. We have reached our objective by introducing a new accurate TS forecasting method, FM2I. Hence, we began by testing the extensivity and finally we reach the intensivity too, it is a pleasant astonishing results. Nevertheless, after careful considerations, it is logic that in order to augment the intensivity we have to find completely new methods rather than still elaborating other methods by only combining or hybridizing existing ones, as usually done in practice.        
\end{rem}

\begin{figure}[!h]
\centering
\includegraphics[width=0.5\linewidth]{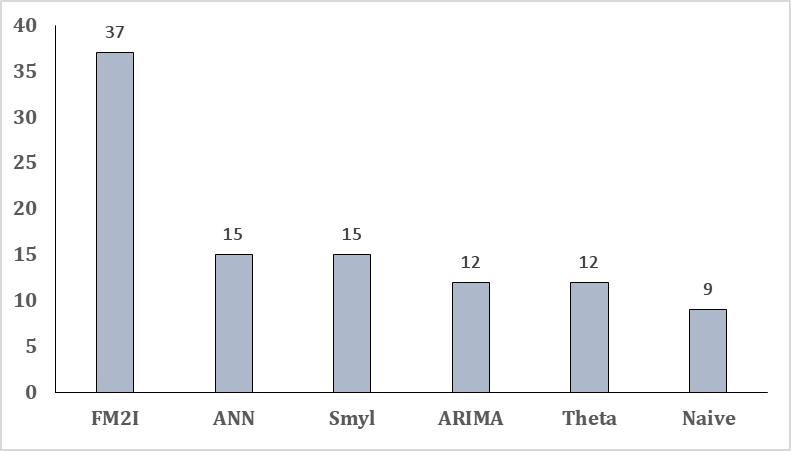}
\caption{The number of TS in which the methods show the highest accuracy:short-term horizon (6) for 100 TS}
\label{M4Fig}
\end{figure}

\begin{rem}[M-Competition]
We have also compared our proposed FM2I against the top methods using the dataset from M4 competition. The dataset contains 100,000 time series, which is categorized into demographics, economic, industry and finance domains \cite{69}. We have selected a significant sample from those available M4 forecasting results. The obtained preliminary results show that the FM2I is ranked the best in terms of accuracy (see Fig. \ref{M4Fig} and Table \ref{M4Tab}). Final testing results will be included in the published version of this work. It is worth noting that the M5 competition is already finished but no results are available so far. The FM2I will be also evaluated against the top M5 methods. Furthermore, we are willing to participate, with the FM2I, in the forthcoming competitions.
\end{rem}

\begin{table}[!h]
\centering
\scriptsize
\begin{tabular}{|l|r|r|r|r|r|r|r|r|r|r|}
\hline
\multicolumn{1}{|c|}{\textbf{Method}} & \multicolumn{5}{c|}{\textbf{Average errors}} & \multicolumn{5}{c|}{\textbf{Rank across all methods}}  \\
\hline
                                      & MSE $.10^3$     & RMSE    & MAE     & MAPE$ \%$  & sMAPE $ \%$ & MSE & RMSE & MAE & MAPE$ \%$ & sMAPE  $ \%$              \\
\hline
FM2I                                  & 1102.07  & 650.4   & 562.49  & 9.37  & 7.91  & 1    & 1     & 1    & 1     & 1                        \\
\hline
Smyl                                  & 2545.43  & 960.21  & 847.53  & 13    & 12.28 & 2    & 2     & 2    & 4     & 2                        \\
\hline
Theta                                 & 3547     & 1033.63 & 912.21  & 12.09 & 12.73 & 3    & 3     & 3    & 2     & 3                        \\
\hline
Naive                                 & 3958.41  & 1118.07 & 984.38  & 12.44 & 13.49 & 4    & 4     & 4    & 3     & 4                        \\
\hline
ANN                                   & 6450.3   & 1379.16 & 1293.3  & 17.36 & 16.93 & 5    & 5     & 5    & 5     & 5                        \\
\hline
ARIMA                                 & 14151.88 & 1458.77 & 1300.24 & 18.95 & 17.69 & 6    & 6     & 6    & 6     & 6                        \\
\hline
\end{tabular}
\caption{FM2I rank against other methods from M4 competition, short-term horizon (6) for 100 TS}
\label{M4Tab}
\end{table}

\section{Related work}
The rapid growth and deployment of IoT (Internet of Things) and wireless sensors technologies have shown a great potential for collecting high-volume, high-velocity, and high-variety of time series data. In fact, a myriad of sensors can be deployed for gathering contextual data that could be integrated with other data, such as location, weather data, and social media data \cite{46}. The processing and analysis of these times series data allows the development of context-aware applications and services in many applications domains, such as in e-health \cite{66}, transportation \cite{65}, and energy management \cite{47}. For instance, short-term forecasting of solar power production and utility demand could allow dynamic and predictive control of micro-grid energy systems \cite{48}. 

There are two main goals in analyzing and processing time series data, classification and regression. Time series data classification allows revealing patterns and features (e.g., anomalous values), while regression allows forecasting/predicting future values according to past previous ones. Time series forecasting and analysis are key elements in many applications. Their aim is to analyze time-series trends by building a forecasting model for being used for either classification or for predicting/forecasting n-steps-ahead values.

\begin{figure}[!h]
\begin{center}
\includegraphics[width=1\linewidth]{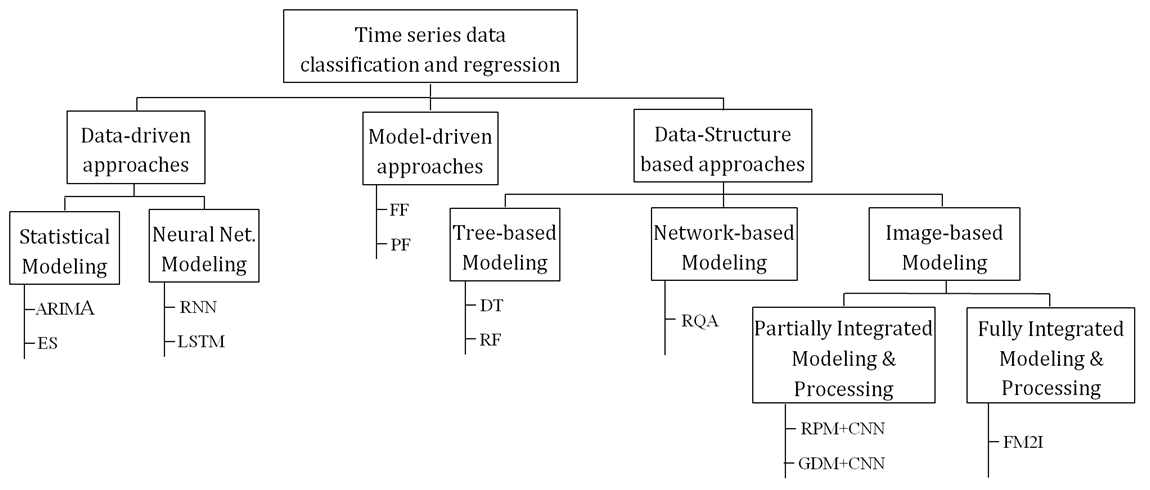}
\caption{Classification of approaches for time series classification and regression}
\label{class}
\end{center}
\end{figure}

Several approaches have been proposed in the past decades for time series data analysis and forecasting. They can be classified into three main categories, as depicted in Fig. \ref{class}: \textit{model-driven}, \textit{data-driven}, and \textit{data structures-driven} approaches. In the model-based approach, time-series are transformed into state-space representations. The forecasting process is then performed by simply assuming that the forecast period observations are missing and hence apply existing methods to complete them \cite{49}. Kalman Filter and Particle Filter \cite{50} are among the model-based methods that can be applied to reveal repeated patterns and for forecasting time series data \cite{51, 52}. While these approaches provide very good accuracy, they remain insufficient in catching the complexity and the dynamics of complex systems. 

Data-driven approaches have been proposed to circumvent this issue by exploiting the high data volumes of measured data. The aim is to establish a more complete data model, which can be used for predicting and forecasting \cite{16}. These approaches can be classified, in turn, into two main categories: \textit{statistical-based} and \textit{Neural Network-based} approaches. Statistical-based approaches have been proposed to predict future values as the product of several past observations. Examples of these approaches are ARIMA \cite{53} and exponential smoothing (ES) \cite{49}. Artificial Neural Network-based approaches have been proposed as information processing models for time series data analysis and prediction \cite{54}. Examples are multi-layer feed forward neural network and LSTM (Long Short Term Memory). However, while these models did provide accurate forecasting results, they typically require larger training datasets in order to learn accurate models \cite{52}.

Data structures-based approaches have been proposed for time series data analysis and forecasting. These approaches can be classified into three main categories: \textit{Tree-based}, \textit{Network-based} and \textit{Image-based} approaches. Classification and regression trees, random forests and gradient boosting trees are the main Tree-based methods that have been proposed for both classification and regression problems \cite{54}. Network-based concepts have been proposed recently for studying and analyzing time series. The aim is to extract and reveal dynamically relevant statistical properties of time series \cite{55}.

Recently, many research studies propose to transform and encode time series into different types of images, such as GASF/GADF (Gramian Angular Summation/Difference Fields) and MTF (Markov Transition Fields) \cite{56}. These approaches fall into partially integrated modeling and processing categories, as described in Fig. \ref{class}. The main aim of most of these studies is to enable and use techniques from image recognition and signal processing communities for time-series classification. All approaches use Artificial Neural Networks in order to reveal topological properties from time series-based images. For instance, in \cite{56}, Wang and Oates propose to use deep Tiled Convolutional Neural Networks in order to classify time series for multi-level features extraction after encoding time series as GASF/GADF/MTF images. Similarly, Yang et al. in \cite{57} have introduced a framework, which allows encoding multivariate time series data into 2D GASF/GADF/MTF-based images, for features classification through a Convolutional Neural Network. The experimental results, using two multivariate datasets, show that these image-based structures do not have a significant impact on the accuracy. Furthermore, Chen and Shi in \cite{58} propose to encode univariate time series data into a Relative Position Matrix and feed it to a Convolutional Neural Network for classification purposes. Unlike the time series classification, similar research studies have been conducted for time series forecasting. For instance, Li et al., in \cite{59}, propose an approach that transforms time series data into graphical representations (in the form of 2D plotted images) and then apply deep learning methods for data forecasting. Similarly, authors in \cite{63} introduce an approach for extracting time series features using computer vision techniques (e.g., spatial bag-of-features \cite{67}, deep CNN). Alike the approach proposed in \cite{59}, authors propsose to first transform time series into graphical representations (recurrence plots) and then use computer vision techniques to extract time series features and reveal the characteristics for being used by the CNN framework for time series forecasting.

However, despite the combination of these image-based structures with state-of-the-art deep learning methods provide the best accuracy, the relationships between extracted topological features and time series data remains unclear, since no inverse operations have been proposed to transform the produced and enhanced 2D images into the new 1D sequence of time series data \cite{56}. Furthermore, most of the approaches proposed so far use partially integrated image-based modeling and processing techniques combined with conventional neuronal approaches as depicted in Fig. \ref{class}.

In the proposed FM2I, a fully integrated modeling and processing framework, as depicted in Fig. \ref{class}, we transform the time series forecasting into fully images- and signal-based processing procedures. In fact, forecasting time series data is transformed into a problem of completing a picture (i.e., without falling back on machine learning techniques). After transforming a time series data into its corresponding image, the problem of \textit{data forecasting} becomes essentially a problem of \textit{image inpainting} problem, i.e., completing the missing pixel in the image. In summary, the main contributions of this work against the above-mentioned approaches are five folds:
\begin{itemize}
\item New matrices are proposed for 2D image-based representations of time series data,
\item New transformation techniques, from time series data to image-based representation,
\item Image-based forecasting techniques based on inpainting methods,
\item Re-inverse transformation techniques, from 2D representation to 1D vector representing the original time series data.
\item Performance evaluation using datasets from M3 competition together with extensive comparison of the FM2I against the top M3 approaches. 
\end{itemize}

It is worth noting that the work presented in this paper initiates a new branch of TS forecasting (see Fig. \ref{class}). In fact, instead of learning TS patterns and features directly from temporal dimension, the spatial representation is used along with image-based processing techniques and tools. More precisely, a 2D image representation could reveal interesting and richer patterns and features, which could not be explicitly represented in a 1D TS. 

\section{Philosophical discussion and other potential directions}
In this work, we have started by pointing out philosophical considerations and reached unexpected and astonishing results: \textit{completing a painting enables us to see the future in color}! A 'visual' 2D predictive procedure works for a 1D chronological data! Even after completing this work, we are not yet able to answer completely the central question concerning the nature of predictions developed in section 2. Nevertheless, we would like to draw some philosophical conclusions from this fruitful endeavor. It is obvious that we need more than a discussion section and certainly more than an article to present in details all our reflections, which we will surely do in forthcoming works. But, we want at least to summarize some of our queries about the implications of this work. More precisely, we have chosen to limit focus on the two following aspects. 

\subsection*{The value of prediction vs explanation: the new emerging role of prediction: Bridging philosophy}

As mentioned in \cite{5}, in contrast with scientific explanation, the value or the role of scientific prediction seems to have always been clear. Indeed, common sense has it that predictions are used for their own interest: \textit{predict} what will be going on in order to make better decisions. At the same time, it helps testing the strength of the efficient way of understanding the surrounding world. In addition, as indicated in Section 2, according to explicit or tacit scientific practices, philosophers of science wish to formalize other interests of predictions as validating or invalidating theories and models and then assessing the maturity of theories (\cite{2b} and \cite{2}). When the value of explanation is unclear as indicated in \cite{5} with reference to the work \cite{4} entitled "Why Ask Why?", see also \cite{28}. More precisely, the argument is: suppose that we possess a perfect power in predicting everything in this world (like Laplace's demon), then the need of explanation becomes unnecessary or serves at most to assuage our thirst for understanding and therefore should be seen as a simple 'psychological satisfaction' which makes it epistemically suspect \cite{29}. The conclusion from this thought-experiment can be corroborated by the actual success of \textit{Data Sciences} based on \textit{Statistics} and \textit{Machine Learning}, which are \textit{black boxes} with explanation free knowledge.

Now suppose we hold the reverse of the argument: we have a perfect explanation power making us capable to explain everything, do we always need predictions? Certainly, yes. Indeed, we need them in order to make, accordingly, the most adequate decisions. At this stage, we point out a certain \textit{Dissymmetry} between predictions and explanations as the value of the first one appears better compared to the second one. There is a bias in this reasoning, though can we considere an explanation without prediction or the reverse? A development concerning the first part of this question can be found in \cite{5} and \cite{29}. More precisely, authors highlight that we can talk about scientific explanations \textit{if and only if} this last is able to generate new testable predictions in order to explore theories implications, which would probably justify why scientists appreciate explanations. This ties in with, in some way, the famous Karl Popper's idea of \textit{falsifiability} or \textit{refutability}. When predictions can arise from explanations but don't necessarily need them (see the definition of pediction considered in Section 2), then the dissymmetry persists. In fact, in the absence of perfect predictive processes, explanations are still needed to properly assess the accuracy of predictions and to accept them by scientists \cite{30}. All in all, we think that as our predictive power grows, so does the need of explanation decrease.

In light of the conclusion of this work, we can point out new perspectives on the role of scientific prediction. Indeed, as depicted in the general framework (Fig. \ref{Fig1}), the bottom line of our idea is to test the extensivity of the theories' or models' predictive capacity. We could actually conclude that: on the one hand, achieving our difficult challenge by succeeding our first test proves that it is doable, and that all without necessarily being experts in the fields studied. On the other hand, it is worth a try it, since this sort of \textit{modus operandi} offers scientists the opportunity to create \textit{Bridges} between different scientific fields/subfields or different scientific theories by following predictions success. At that moment of our experiments, we wonder if there are other examples of these bridges in science. After carreful and extensive research, we found some examples of the tacit applications of the Fig. \ref{Fig1}. We chose to present a significant example from theoretical physics, which also focuses the attention of some philosophers of science \cite{34} and \cite{35}, but never from the angle developed here. Thus, let us mention the famous so called \textit{AdS/CFT Correspondence}, first introduced by Juan Maldacena in the mostly cited article of high energy physics (20000 citations) \cite{31} and most significant achievement in string theory in the last twenty years. From our perspective, this example illustrates perfectly our ideal conclusion.

\begin{figure}[!h]
\centering
\begin{subfigure}{.48\textwidth}
  \centering
  \includegraphics[width=1\linewidth]{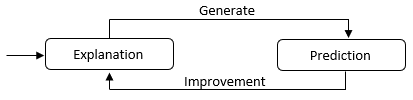}
  \caption{}
  \label{CL1}
\end{subfigure}%
\hspace{0.5cm}
\begin{subfigure}{.48\textwidth}
  \centering
  \includegraphics[width=1\linewidth]{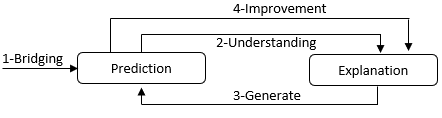}
  \caption{}
  \label{CL2}
\end{subfigure}

\caption{Closed loop: a) conventional, b) new} 
\label{MAT}
\end{figure}

Indeed, in theoretical physics, the anti-de Sitter/conformal field claims the equivalence between the strongly coupled four dimensional gauge theory and gravitational theory in five dimensions. In particular, it can be seen as a bridge between two completely different fields of physical theory, since it is an unexpected link between Gravity theory in five dimensions and Quantum field theory in four dimensions. This correspondence deserved to be explained in details, for the sake of conciseness, we refer the interested readers to \cite{33} and \cite{31}. Nonetheless, we must point out that there is no formal mathematical proof of this correspondence, even if, there exists an \textit{AdS/CFT dictionary} permitting to switch between the two theories. An example of applying the previous process can be clearly identified in nuclear physics. Briefly, in order to study quark-gluon plasma, instead of using its insoluble mathematical formalism, in \cite{32}, according to the AdS/CFT correspondence, authors predict the value of the ratio of shear viscosity to the entropy density associated with the quark-gluon plasma, which is close to the experimental results. This leads to a new understanding and then new explanations concerning quark-gluon plasma through this correspondence. Here, we reach a new benefit of prediction since using this precess we may find, if even possible, new explanations by reversing the conventional closed loop, depicted in Fig. \ref{CL1}.

Starting by testing the extensivity of theories or models' predictive capacity, we select the theory or model, which provides the most accurate predictions (i.e., there is a bridge) and only after we go back to the previous process by trying to obtain explanations and so on (see Fig. \ref{CL2}).

In fact, by systematizing the approach, described in this work, and by identifying the possible bridges according to their predictions accuracy, we will surely find more correspondences, and should not be dependent on the scientists’ genius like Maldacena. In the next subsection, we will try to go further by investigating the deeper reason for the existence of these bridges. Nevertheless, this interesting gain relies on overcoming the major difficulty of how we can switch between scientific disciplines/sub-disciplines. More precisely, it is worth experimenting how we can transform initial conditions or data's field/subfield to be an input of a theory or a model of another one? In other words, we wish to develop in forthcoming works a more philosophical/mathematical framework, called \textit{Correspondence}, when we hope to formalize in details this idea (we have very promising leads). In our example (FM2I), we have moved from 1D temporal series data structure to a suitable 2D image data structure. First of all, let us begin by designating the benefits of these bridges by giving three essential aspects, which can also be considered as novel roles of scientific predictions:    
\begin{itemize}
\item \textit{New models and methods}: prediction has always helped scientists to select between models and methods inside a field/subfield. But, in the light of this work, it appears that this selection could be generalized between models and methods from other fields/subfields, since we found a correspondence permitting us to switch between them. In our example, we have found new forecasting models and methods arising from image inpainting methods.      

\item \textit{New Correspondences and Analogies}: we will develop our perspective in details concerning these notions in a forthcoming work. So far, we have presented above an example of correspondence based on a dictionary permitting to switch between two theories. We can also obtain other types of correspondence through other transformations like those developed in this paper. Of course, the different types of correspondences should be chosen according to their predictions accuracy. Here, we meet another new role of prediction. Analogies are also a type of correspondence, which are very fruitful processes used by many greatest scientist like J.C. Maxwell and H. Poincar\'e (see \cite{36} and \cite{37}). Then, we hope we can find new analogies following this procedure.

\item \textit{Towards new theories}: the correspondences selected by predictions could lead to new theories, a discussion concerning an example related to analogies between \textit{Heat and Electricity} can be found in \cite{38}.  

\end{itemize}

\subsection*{Why do these bridges exist? Extensive Structural Realism}

Wondering the reason of these bridges' existence, we have naturally converged towards a modified version of the Poincar\'e's \textit{Structural Realism} thesis. Let us begin by briefly recalling some philosophy of scientific modes of thinking in order to better clarify our position. Concretely, \textit{Scientific Realism} posits that mature theories with an undisputed predictive success are true or approximatively true in the sense that the structure assumed in these theories reflect the actual reality. The main argument in support of scientific realism is called the \textit{Miracle Argument}, which is introduced by Hilary Putnam when he wrote \textit{"The positive argument for realism is that it is the only philosophy that doesn't make the success of science a miracle"}\cite{39}. This means that it is virtually impossible that our theories succeed to describe observable phenomena and at the same time be completely false. The principal argument against scientific realism is called \textit{Pessimistic Induction}. Based on history of science, this theory claims that since most of our past theories, even strongly confirmed, have turned out to be incorrect, thus, it is very improbable that our current theories be true. According to this argument, the present theories are ephemeral and are likely to change; it is just a matter of time.

In an article entitled \textit{"Structural Realism: The Best of Both Worlds?"} \cite{40}, John Worrall elaborated a position inspired by Poincar\'e's works. More precisely, structural realism takes into the account the pessimistic induction concerning the theoretical entities involved in current science, but at the same time it maintains the validity of the miracle argument at the mathematical structures level. Indeed, the famous example proposed in \cite{40} and also analyzed by Poincar\'e in \cite{41} is very relevant. Specifically, Fresnel's wave theory is based on the wrong postulate of an entity existence, called \textit{Luminiferous Aether}, as the medium for the propagation of light. Augustin-Jean Fresnel developed a mathematical formula, called \textit{Fresnel diffraction integral}, which was validated since it succeeded in observing and predicting the famous Frenel's spot. This example proves that there is some truth in Frenel's theory, which is clearly its mathematical structure and not the existence of an entity permitting the propagation of light. In fact, Frenel's equations can be seen as a particular case of Maxwell's equations, which does not assume any ether existence. Therefore, Frenel's mathematical structure can be viewed as approximatively true since it is conserved as a particular case in the later theories dealing with light propagation like electromagnetism, including quantum theory.

In this respect, we think that we should not be confined to conserve mathematical structures in later theories treating the same phenomenon. We have to distinguish between \textit{Intensive} and \textit{Extensive} structural realism. Indeed, we can posit that this propriety of conserved mathematical structure is an \textit{Intensive Structural Realism}. While, in the light of this article development, we are able to introduce what an \textit{Extensive Structural Realism} can be. Specifically, in most cases, testing the extensivity of theories' predictive capacity means testing the extensivity of the predictive capacity of its mathematical structures. Indeed, the above example of the prediction concerning the quark-glon plasma according to the AdS/CFT correspondence is specifically due to the mathematical language of the string theory. More accurately, we can conclude that the capacity of certain mathematical structures of working in several domains means that there exists a kind of universality of these structures. This allows us to think about the miracle argument at the level of the mathematical structures extensivity, in the sense that it should be a miracle that a mathematical equation or formula appears in many completely different domains without reaching any reality. We can find many examples (see \cite{36}) in physical analogies when an equation or formula works in different fields. In this sense, extensive structural realism is the capacity for some mathematical structures to arise in domains other than those for which it was initially elaborated.  

Let us provide another example in relationship with this work. We are currently developing a forecasting promising method based on some partial differential equations like Heat and Navier-Stokes equations through adequate bridges (image). Knowing that heat equation, for example, was developed by Joseph Fourier to model heat diffusion; nevertheless, this equation (or its modified version) appears in many other domains like electricity, image restoration, Brownian motion, finance, quantum mechanics, forecasting, etc, which means that this equation possesses something universal, at least at a first order. Thus, it should be a miracle that this equation be completely divorced from a certain reality. That is what we called the \textit{extensive structural realism}. Hence, as one of the objectives of this work is to highlight the interest of creating new bridges between theories and models we naturally highlighted the question of their existence. After careful consideration, we strongly think that this existence is due to the existence of mathematical structures, like heat equation; in other words, the miracle argument at the level of the mathematical structures recurrence or persistence. The basic idea is then, is that we are sure that these types of mathematical structures exist, since we possess many examples, then we have to strengthen the existing and find new ones by trying to test bridges according to their predictions' accuracy.

We must point out that the discussion above suggests a significant link with some modes of thinking from \textit{Philosophy of Mathematics}. For example, it should be very interesting to further analyze these ideas in light of the \textit{Mathematical Realism} concerning the question of mathematics and mathematical objects' objectivity. We will focus on such links in a forthcoming work. Let us conclude this subsection by the following reflection. Elaborating a theory of everything, capable of describing and explaining all phenomena in the universe, is a precious physicists' hope. One of the main impediments is the difficulty of finding a theory unifying gravitation and standard model of particle physics. Whether this unification is really possible, it is reasonable to think that it should be realized through a mathematical structure. Hence, this last must be able to explain and predict everything. We think that it will be more suitable to proceed by trying to find this structure through its capacity of predicting everything and only after that check (or not) its capacity of explaining and describing everything, and not the reverse, which seems much more difficult. Indeed, if we are able to evacuate the major difficulty of bridging theories and models through adequate correspondences, we will be able to test this procedure. Who knows? Perhaps the equation of everything already exists in the literature; it just needs to be tested!   

Let us conclude this section by the following remark.

\begin{rem}[Philosophy is not dead]

In \cite{42}, the famous theoretical physicists Stephen Hawking and Leonard Mlodinow wrote: \textit{"Philosophy is dead"}. This statement naturally made many philosophers react. Without wishing to engage in polemics, this article is a direct refutation of this declaration. Indeed, it is well known in logic that a counterexample is better than any argument. This article is clearly a counterexample since it proves the power of philosophy. More precisely, we are neither forecasting specialists nor signal processing specialists; nevertheless, starting by philosophical considerations and using our scientific background, we have created a new accurate forecasting method by bridging two domains. In the light of this work, we strongly think that experimental philosophy of science is an interesting and promising way combining philosophy and science capable to create new and innovative research fields.

\end{rem}

\section{Scientific discussion and other potential directions}
The ideas developed in this article are clearly the beginning of a large series of works since there is an important variety of other potential directions. We briefly focus on two aspects or questions: is the FM2I improvable? Why it works? Should it be adapted for real-time data streaming?        

\subsection*{Parameters tuning and optimization}

\begin{figure}[!h]
\begin{center}
\includegraphics[width=0.5\linewidth]{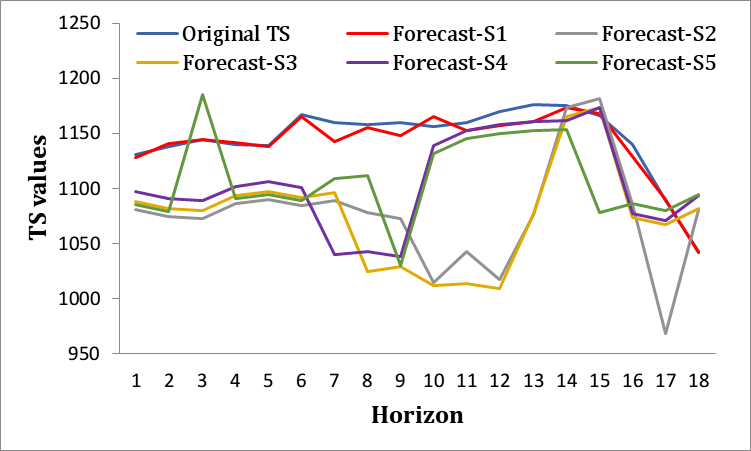}
\caption{Forecasting values: Forecast-S1 represents the best fitting model}
\label{improvFig}
\end{center}
\end{figure}

The parameters tuning of the FM2I forecasting algorithm is based on a grid search through testing numerous combinations of matrices (conventional and differenced ones), min-max TS scaling and patch size. It generates a set of best candidate forecasting models during the progressive TS exploration. The aim is to extract the most frequent and best suitable configuration. Generally, this model provides accurate forecasts, however, our exhaustive forecasting approaches shows that this model is not necessarily the best fitting one. Fig. \ref{improvFig} and Table \ref{improvTab} show, as an initial proof of this statement, the improvement of FM2I when using the best fitting model. The accuracy differences show that adequate improvement could be performed in order to select the best model (i.e., Forecast-S1 in the current case, near the real TS). This could be achived through adapted and well known optimization approaches.

\begin{table}
\centering
\scriptsize
\begin{tabular}{|l|r|r|r|r|r|}
\hline
\multicolumn{1}{|c|}{\textbf{Scenarios}} & \multicolumn{1}{c|}{\textbf{MSE}} & \multicolumn{1}{c|}{\textbf{RMSE}} & \multicolumn{1}{c|}{\textbf{MAE}} & \multicolumn{1}{c|}{\textbf{MAPE }} & \multicolumn{1}{c|}{\textbf{sMAPE }}  \\
\hline
Forecast-S1                          & 61,44                             & 7,83                               & 5,61                              & 0,48\%                              & 0,49\%                                \\
\hline
Forecast-S2                                       & 7161,03                           & 84,62                              & 74,93                             & 6,54\%                              & 6,81\%                                \\
\hline
Forecast-S3                                     & 7886,58                           & 88,80                              & 74,86                             & 6,50\%                              & 6,80\%                                \\
\hline
Forecast-S4                                     & 3537,93                           & 59,48                              & 46,42                             & 4,06\%                              & 4,18\%                                \\
\hline
Forecast-S5                                      & 3043,92                           & 55,17                              & 47,21                             & 4,12\%                              & 4,22\%                                \\
\hline
\end{tabular}
\caption{Metrics for the best scenario via the best fitting model: Forecast-S1}
\label{improvTab}
\end{table}

\subsection*{Towards ensemble data autocorelation forecasting}

Several methods, as depicted in Fig. \ref{class}, have been proposed for TS forecasting. Many recent studies have shown that none of these forecasting methods is alone able to model real-world TS. Hybrid ensemble models, named also ensemble learning model, have been proposed for accurate TS forecasting, combining then strengths of each forecasting method. For instance, they have been heavily used in the past decades for weather forecasting \cite{79}. These models' aim is to combine several forecasting models in order to increase the accuracy against the usage of a single model \cite{70}, \cite{72}. In fact, an ensemble of independent models (i.e., multiple and diverse) could be established for being used in the collective prediction process\cite{73}. For instance, in \cite{71} authors propose a hybrid model, combining ARIMA and Elman artificial neural network (EANN), for TS forecasting. Reported results show that the proposed hybrid model outperforms both ARIMA and EANN when considered separately. Similarly, authors in \cite{75} propose an ensemble learning model composed of the decision tree, gradient boosted trees and random forest models and results show the effectiveness of this hybrid approach. Most of work to-date show that combining multiple forecasts, which are generated by an ensemble models, could generally provide better forecasts with higher accuracy \cite{76}. However, their efficiency was effectively highlighted in the M4 competition by the Smyl's approach, which won the competition. This later, named ensemble of specialists, combines the LSTM (long short term memory) neural network with an ES model (exponential smoothing) (ES) model, for TS forecasting. 

Unlike ensemble learning models, in our poposed FM2I, we are able to forecast $O(n^{h})$ possibilities, where $n$ is the TS size and $h$ is the considered horizon. Thus, FM2I can be seen as \textit{ensemble data autocorrelation forecasting (EDAF)} in which, instead of generating an ensemble of learning models, several ensemble data are generated (i.e., forecast scenarios). This later is then used to select the best suitable forecast that maximizes the accuracy.

\subsection*{Augmented dimension prediction and entropy reduction}

Revisiting this works' outcome, we naturally put more emphasize on the following question: Why does this bridge works so well? The first idea that we have highlighted is: it should be a link with augmenting dimension! Indeed, we start by wishing to predict forthcoming values of 1D TS, we transform it to a richer 2D structure and then predict a region related to these values. It turns out that this forecasting method is very accurate, and then we test the effect of augmenting dimension. We will explain in details this link in a forthcoming work. The second idea we are investigating concerns the entropy. Indeed, it is well known that the entropy is related to TS's complexity and then to the forecasting performance \cite{77} \cite{78}. Thus, basically, according to the accuracy of FM2I, we have absolutely augmented the \textit{forecastability} of our initial TS. This indicates that we have reduced the initial entropy (somehow) of TS by augmenting its information through our TS-image transformations. In other words, the idea is to measure the amount of information gained for each TS-image transformations.             

\subsection*{Real-time TS forecasting}
Real-time TS streams processing induces two main challenges, limited resources and frequent data distribution changes. The first challenge requires using distributed computing platforms for handling the high volume and velocity of data streams. Regarding the second challenge, time series are usually received from data sources with several imperfections, such as noise and/or redundancies, missing values and inconsistencies. Consequently, predictive algorithms will perform poorly using low quality data. Data preprocessing or data preparation techniques, such as cleaning integration, normalization and transformation are highly required in order to deal with this issue. In addition, the amount and the dimension of received data is growing considerably with the emergence of IoT devises will be connected and embedded in our environments (e.g., buildings, vehicles), and reduction techniques become mandatory for dimensions reduction and simplification (e.g., feature selection). These techniques will help in providing high quality data with reduced dimensions, which are necessary for faster training and better understandability of results. Another important characteristic of data is their continuous and high speed arrivals (i.e., data streams). Many emerging real-world applications are generating data streams, and therefore, dataset ever-grow inconsiderably. However, techniques are required to cope with the time and memory constraints of the high arrivals of data streams. These streams might have a non-stationary behavior that could lead to the concept drift in which the distribution of data streams change frequently \cite{64}. So, learning techniques must be adaptive and learn from new arrival data (i.e., adapt to changes in the processes that generate data streams) especially for context-driven applications requiring real-time or near time decisions (e.g., road accident avoidance, patient monitoring). In this direction, FM2I is under deployment in real-sitting scenarios in order to study its effectiveness in dealing with real-time TS streams. We are also adapting it to deal with multivariate TS for both batch and stream processing.

\section*{Acknowledgments}
Authors sincerely thank Prof. Najib MOKHTARI for the proofreading of the paper. We would like to thank Prof. Bertrand Denise. Indeed, looking for a significant example from physics to support our concept, we have explained our philosophy of bridging and founding correspondences, he then naturally provide us the example of AdS/CFT correspondence. Special thanks to Prof. Alain Degiovanni for his encouragement, in particular, he always believe in this idea.

\section*{Appendix}
\subsection*{Appendix A}
Generally, data are transmitted as electrical signals. Thus, by electrical analogy, signals could also be classified according to their energy characteristics, signals with finite energy or with finite power. In our case, we consider a signal $x(t)$ as a random with a finite average power. This type of signals is also called a permanent signal. In other words, a signal $x(t)$ is permanent if 
$P(x(t))=\lim_{T\to +\infty}\frac{1}{T}\int^{\frac{T}{2}}_{-\frac{T}{2}} |x(t)|^2dt<+\infty$. Actually, it is practical and reasonable to add the condition of being bounded. In other words, we will consider finite power signals, which are in addition an element of $L^{\infty}(\mathbb{R})=\{x(t)/\sup_{t\in\mathbb{R}}|x(t)|<+\infty\}$. The class of functions $L^{\infty}(\mathbb{R})$ is a smaller than the class of functions with finite average power because if $x\in L^{\infty}(\mathbb{R})$ then it is a finite power:
$$\lim_{T\to +\infty}\frac{1}{T}\int^{\frac{T}{2}}_{-\frac{T}{2}} |x(t)|^2dt\leq \sup_{t\in\mathbb{R}}|x(t)|^2=\Vert x \Vert^2_{L^{\infty}(\mathbb{R})} <+\infty$$
The class of finite average power signals is a fairly broad class because it also contains finite energy signals. Indeed, a signal $y(t)$ is said to be of finite energy, also called transient signal if
$$W(y(t))=\int_{\mathbb{R}}|y(t)|^2dt<+\infty\Leftrightarrow \Vert y\Vert_{L^2(\mathbb{R})}<+\infty$$. 

Thus, the finite energy signals are the functions of squares summable on $ \mathbb{R}$. In addition, a signal with a finite energy is a signal with zero average power and, therefore, is finite. Indeed, we have:
$$P(y(t))=\lim_{T\to +\infty}\frac{1}{T}\int^{\frac{T}{2}}_{-\frac{T}{2}} |y(t)|^2dt\leq \lim_{T\to +\infty}\frac{1}{T}W(y(t))=0$$.

In practice, signals are given over a finite duration and their average power $P$ is, therefore, calculated over a window of finite size. This implies that signals with finite energy will have a small but not zero average power. For this reason, we have considered a large class, which includes practically all varieties of random signals. Furthermore, the average power and the energy can be calculated in the discrete case, namely, for a digital signal. So, to summarize, we can consider our time series as digital signals resulting from the sampling of permanent random and analog signals $x(t)$ with finite and bounded power.

\subsection*{Appendix B}
The PSD allows specifying the frequency content of a signal. It is important to note that the spectrum of a signal $x(t)$ is the modulus of its Fourier transform $|\mathcal{F}(x(t))|=|X(\xi)|$, where $\mathcal{F}$ denotes the Fourier transform. We then define the PSD of the signal $x(t)$ as the square of the spectrum of $x(t)$, namely, $S(\xi)=|X(\xi)|^2$. In our case, the signal $x(t)$ is assumed to be at finite average power. It is, therefore, not necessarily integrable over $\mathbb{R}$ and its Fourier transform does not necessarily exist, at least in this form. We can, however, obtain the same formula asymptotically. More precisely, we can define the Fourier transform on a window of size $T$, denoted by $X(\xi,T)$. Thus, we can speak of spectral density on a window, which is named by $S(\xi,T)=|X(\xi,T)|^2$. Therefore, the PSD of this type of signal can be defined by
$S(\xi)=\lim_{T\to +\infty}\frac{1}{T}|X(\xi,T)|^2.$

Now, we can make explicitly a relationship between $\Gamma(\tau)$ and $S(\xi)$. Indeed, using Parseval's theorem we can show that the PSD is the Fourier transform of the temporal autocorrelation function (here the Fourier transform can be in the sense of a distribution), $S(\xi)=\mathcal{F}(\Gamma(\tau))=\int_{\mathbb{R}}\Gamma(\tau) \exp(-2i\tau\xi)d\tau$.

\subsection*{Appendix C}
So far, we have clarified, from signal processing perspectives, the relationships between time series and signals. We also explained the choice of (STAM) in relation to its properties and highlighted its features from signal processing principles. This part focuses, however, on the particular case dealing with stationary and ergodic random signals. More precisely, we have considered (TS) as a random digital signal, so it can be considered as a discrete stochastic (or random) process indexed by time $(X_i)_{i\in\mathbb{N}}$. It is a family of random variables, at each instant $i\in \mathbb{N}$, the random variable $X_i$ has a certain probability distribution and the value $s_i$ of (TS) is an instance of it. We suppose that all these random variables $(X_i)_{i\in\mathbb{N}}$ have probability densities that we denote by $(f_i)_{i\in\mathbb{N}}$. We also denote by $ f_ {i, j} $ the joint probability density of the two random variables $X_i$ and $X_j$.

Let us denote by $\mathbb{E}(X)$ the expectation of $X$. The \textit{Statistical Autocorrelation Function} between time $i$ and $j$ is written:
$$R(i,j)=\mathbb{E}(X_iX_j)=\int_{\mathbb{R}}\int_{\mathbb{R}}x_ix_jf_{i,j}(x_i,x_j)dx_idx_j.$$
This function is linked to an important function, it is the function of auto-covariance.
Now, if we want to use the tools of frequentist statistics (or classical statistics), we start by assuming, at least, that the stochastic process $(X_i)_{i\in\mathbb{N}}$ is \textit{stationary}.

\begin{defi}
A random process is said to be stationary in the strict/strong sense (or to order 1) if all its statistical properties are invariant at any time scale. More precisely, all the $(X_i)_{i\in\mathbb{N}}$ have the same probability distribution and for any indices $i\neq j$ the joint distribution $f_{i,j}$ is invariant by translation of time, i.e., $f_{i,j}=f_{l,k}$ si $i-j=l-k$. 

The process can be seen as stationary in the broad/weak sense (or of order 2) if the statistical properties of order 1 (expectation and variance) and 2 (statistical autocorrelation function) are invariant over time.
\end{defi}

As strict stationary $\Rightarrow$ stationary broad, if the stochastic process $(X_i)_{i\in\mathbb{N}}$ is stationary strict or broad, then we will have:
$$R(i,j)=R(\tau)\ \ \ avec\ \ \ i-j=\tau,$$ 
$$\forall l\in\mathbb{N}, R(i,j)=\mathbb{E}(X_lX_{l-\tau})\ \ \ avec\ \ \ i-j=\tau.$$

\begin{theo}[Wiener-Khinchin-Einstein theorem]
For a stationary stochastic process in the broad sense, the PSD is the Fourier transform of its statistical autocorrelation function
$$S(\xi)=\mathcal{F}(R(\tau))=\int_{\mathbb{R}}R(\tau) \exp(-2i\tau\xi)d\tau.$$ 
\end{theo}

This result is very important; it has been independently demonstrated by Wiener and Khinchin. Its statement was also anticipated by Einstein. This first theorem gives us a first link between the functions $\Gamma$ and $R$, the temporal and statistical autocorrelation. Indeed, the Fourier transform of the two functions gives the PSD.

\begin{defi}A random process is said to be ergodic if the statistical means are equal to the temporal average values:
$$\mathbb{E}(h(X))=\int_{\mathbb{R}}h(y)f(y)dy=\lim_{T\to +\infty}\frac{1}{T}\int^{\frac{T}{2}}_{-\frac{T}{2}} h(x(t))dt.$$
\end{defi}

Thus, if our process is stationary (wide) and ergodic, we have an equality between the statistical autocorrelation function and the temporal autocorrelation function:
$$\Gamma(\tau)=R(\tau),$$
Thus, our (STAM), defined for any (ST), is equal to the statistical autocorrelation matrix for an ergodic stationary stochastic process. Stationarity is testable but the ergodicity, however, could not be tested, it is only assumed. 

\begin{rem} 
If we use the Wiener-Khinchin-Einstein theorem, as the PSD is the Fourier transform of $\Gamma$ and $R$. Thus, in the cases where the Fourier transform is bijective we will have the following, without any assumption on the ergodicity of the stochastic process. 
$$\mathcal{F}^{-1}(S)(\tau)=\Gamma(\tau)=R(\tau),$$
This is done in particular for $\Gamma$ and $R$ belonging to the Schwartz space or $L^2(\mathbb{R})$. Thus, if we assume that the signals are fairly regular, we can assume that they are ergodic.
\end{rem}

\end{document}